\documentclass[11pt]{article}

\usepackage{acl}
\usepackage{times}
\usepackage{latexsym}

\usepackage[T1]{fontenc}
\usepackage{amsmath,amssymb}
\usepackage{bbm} 
\usepackage{graphicx}
\usepackage{subcaption}
\usepackage{xcolor}
\usepackage{multirow}
\usepackage{algorithm}
\usepackage{listings}
\usepackage{verbatim}
\usepackage{booktabs}
\usepackage{tabularx}
\usepackage{makecell}
\usepackage{float}
\usepackage{algorithmic}
\usepackage{amsmath}
\usepackage{makecell}
\usepackage{tcolorbox}
\usepackage{booktabs}
\usepackage{enumitem}
\usepackage[title]{appendix}
\usepackage{needspace}
\usepackage{placeins}
\tcbuselibrary{breakable}
\lstdefinestyle{mystyle}{
    language=Python,
    basicstyle=\ttfamily\normalsize,
    breaklines=true,
    breakatwhitespace=false,
    showstringspaces=false,
    tabsize=4,
    columns=flexible,
    keepspaces=true,
    frame=single,
    rulecolor=\color{black!30},
    keywordstyle=\bfseries,
    commentstyle=\color{gray},
    stringstyle=\color{blue!40!black},
}
\usepackage[utf8]{inputenc}

\usepackage{microtype}

\usepackage{inconsolata}

\usepackage{graphicx}

\title{EvoCurr: Self-evolving Curriculum with Behavior Code Generation for Complex Decision-making}

\author{
Yang Cheng$^{3}$\thanks{Equal contribution.}%
\thanks{Work done while interning at Zhongguancun Academy.\\
\texttt{chengyg@mail.ustc.edu.cn}} \qquad
Weiyu Ma$^{4}$\footnotemark[1] \qquad
Zilai Wang$^{5}$ \qquad
Wenhui Zhu$^{1,2}$ \qquad
Yujing Hu$^{6}$ \\[4pt]
\bfseries Tangjie Lv$^{6}$ \qquad
Mohamed Elhoseiny$^{4}$ \qquad
Yue Deng$^{1,2}$ \qquad
Jian Zhao$^{1,2}$ \\[10pt]
$^1$ Zhongguancun Academy \\
$^2$ Zhongguancun Institute of Artificial Intelligence \\
$^3$ University of Science and Technology of China \\
$^4$ King Abdullah University of Science and Technology (KAUST) \\
$^5$ Xi'an Jiaotong University \\
$^6$ Fuxi AI Lab, NetEase Inc.
}

\begin{document}
\maketitle

\begin{abstract}
Complex decision-making often requires agents to progress through intermediate tasks rather than solve the final target directly. Existing LLM self-refinement methods typically iterate on a fixed target, while curriculum-learning methods often rely on hand-designed schedules, training-time optimization, or domain-specific difficulty metrics. To smooth out the learning curve with adaptive curriculum desgin, we introduce \textsc{EvoCurr}, a general inference-time framework that co-evolves curricula and executable policies. A Designer proposes verifiable intermediate tasks from the latest accepted progress and recent failures, while a Solver generates or trains policies for these tasks. A task-policy pair is accepted only when it passes hard feasibility checks and reaches a specified performance threshold meanwhile an accepted-floor constraint preserves the latest verified checkpoint and prevents failed harder attempts from overwriting mastered behavior. We evaluate \textsc{EvoCurr} across code-as-policy and closed-loop MARL settings on StarCraft II super-late-game micromanagement, hardest stress test, where \textsc{EvoCurr} achieves 96.7--99.2\% macro-average win rates across three LLM backbones without LLM parameter training. The performance improvments on other experiments, Flatland, POGEMA, Overcooked, and LiveCodeBench Hard, reflect the generality of our EvoCurr.
\end{abstract}

\section{Introduction}
\label{sec:intro}
\begin{figure*}[t]
    \centering
    \includegraphics[trim=1cm 2cm 1cm 1.5cm, clip, width=1\linewidth]{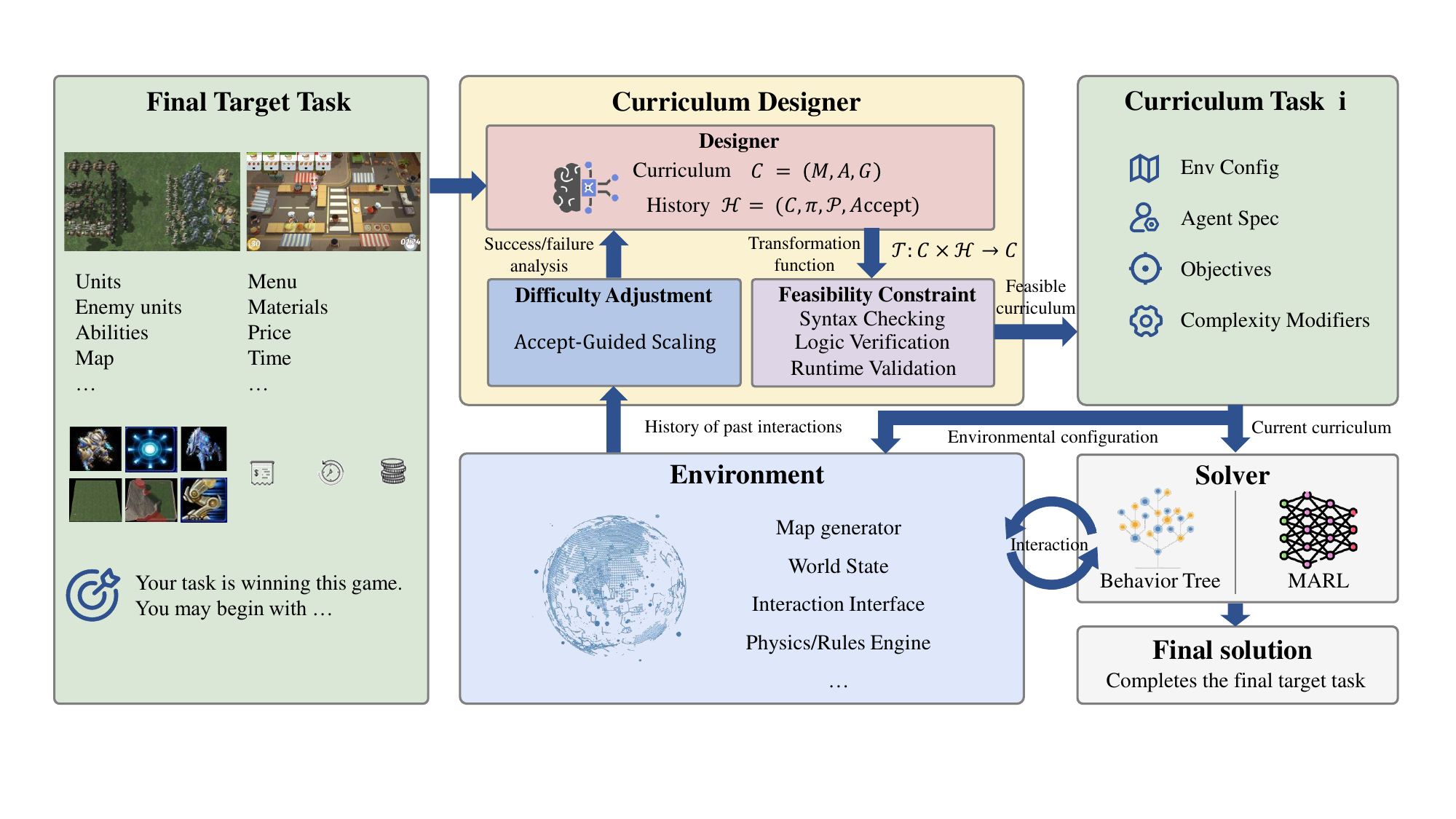}
    \caption{EvoCurr overview. A curriculum designer proposes the next curriculum $C_{t+1}$; a solver produces an executable policy $\pi_{t+1}$ and evaluates it; the outcome feeds back to the designer. The loop starts from a simplified version of the final target $T_f$ and proceeds until $T_f$ is solved.}
    \label{fig:architecture}
\end{figure*}

Complex decision-making tasks are rarely solved by attempting the final objective in one step. Agents often need to master simpler subproblems, preserve the successful behaviors, and gradually extend them toward harder goals. Curriculum learning formalizes this intuition, but practical curricula
are still difficult to deploy broadly, as many methods require
hand-designed task sequences, training-time schedules, or domain-specific
difficulty metrics \citep{bengio2009curriculum, wang2021survey,
soviany2022curriculum}. These assumptions are restrictive for LLM-based agents, which are increasingly expected to solve new tasks at inference time by code generation, interpreting feedback, and interacting with external environments.

Most LLM self-improvement methods also do not explicitly construct
curricula. They usually keep the target fixed: the model generates a
solution, observes feedback, and then revises another solution for the
same final task \citep{shinn2023reflexion}. Such fixed-target refinement can repair local errors, but it is often insufficient when the final hard task combines executable validity, long-horizon control, multi-agent coordination, and environment-specific constraints. In these settings, progress may require changing the task itself. The system should identify intermediate tasks that are solvable, verify the corresponding policies, and use them as stepping stones toward the final objective.

We propose \textsc{EvoCurr}, an inference-time framework for curriculum-policy co-evolution. Given a final task and an acceptance criterion, a Designer proposes the next curriculum task using the final target, the latest accepted progress, and the history of recent successes and failures. A Solver then generates or trains a policy for the proposed task. The candidate is accepted only if it is feasible and reaches the required score. During the evolution process, the easibility is enforced through hard checks such as task-schema validation, code compilation when applicable, and runtime execution. Meanwhile, the accepted-floor constraint enables \textsc{EvoCurr} advance only when a candidate task-policy pair is verified. The latest accepted checkpoint is preserved when the Solver fails a harder task. This separates productive exploration from policy regression and allows the system to search over task difficulty at inference time.

\begin{figure*}[t]
    \centering
    \includegraphics[trim=1.75cm 2cm 1.75cm 2cm, clip, width=1\linewidth]{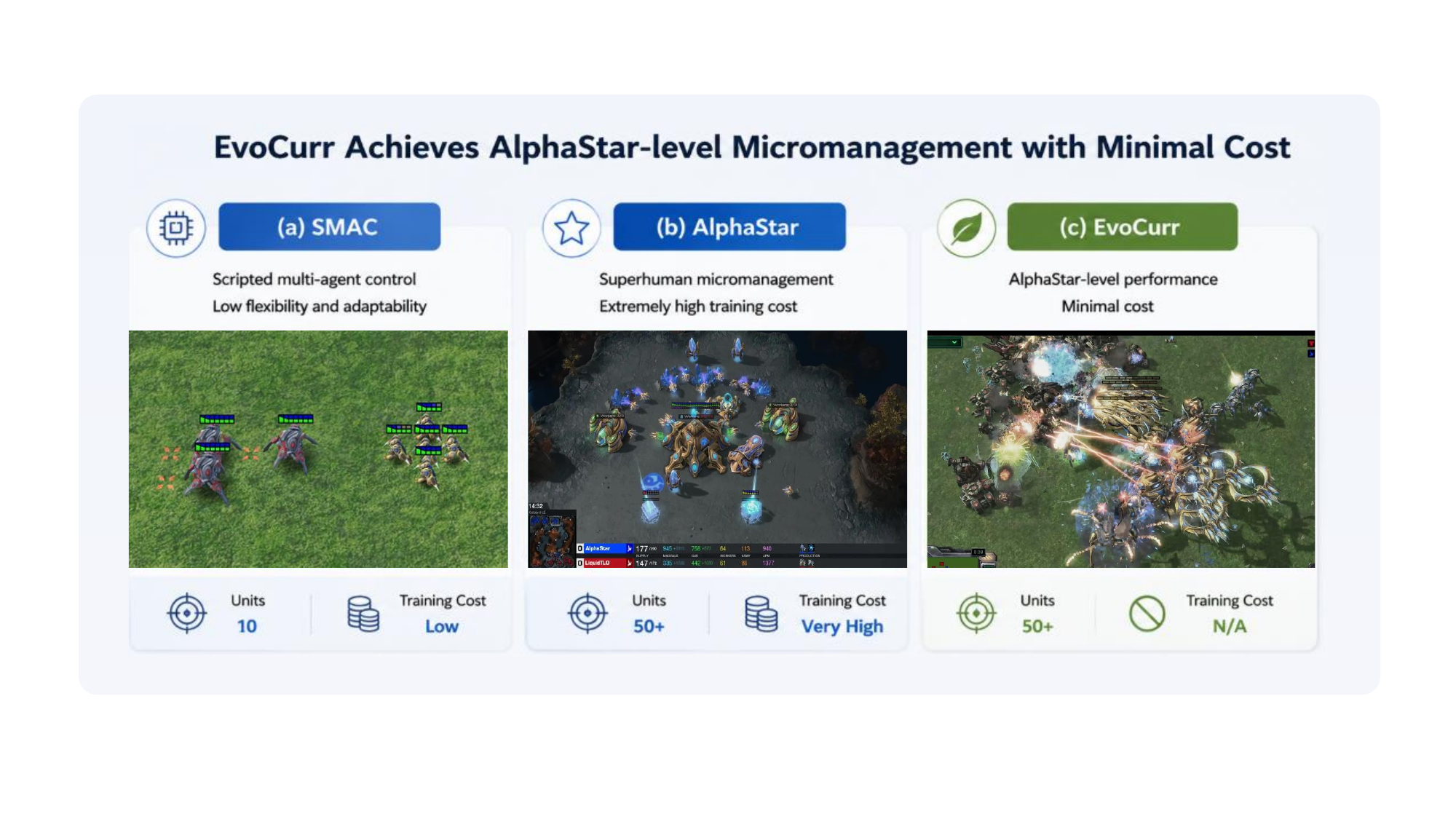}
    \caption{Positioning EvoCurr against existing StarCraft II paradigms. (a) SMAC targets small-scale skirmishes with limited flexibility. (b) AlphaStar handles large-scale, full-game complexity but requires prohibitive training compute. (c) EvoCurr tackles complex, super-late-game micromanagement (50+ units) at inference time, achieving high-level coordination with zero parameter training cost.}
    \label{fig:compare}
\end{figure*}

We use StarCraft~II super-late-game micromanagement as the hardest stress test for this idea. As shown in Figure~\ref{fig:compare}, this setting sits between SMAC-style fixed skirmishes~\citep{samvelyan2019starcraft} and AlphaStar-scale full-game training~\citep{alphastarnature}: economy, scouting, and production are removed, but the remaining battle still requires coordinated control over large heterogeneous armies under simulator feedback. Recent LLM-based StarCraft agents show that language models can reason about strategies and write game-playing code, while also exposing weaknesses in late-game control, long-horizon planning, and policy interpretability~\citep{ma2024largelanguagemodelsplay}. This makes the domain a compact test of whether an LLM can turn local tactical knowledge into one coherent executable controller.

Across $12$ StarCraft~II late-game micro tasks, \textsc{EvoCurr} evolves
interpretable behavior-tree policies with no parameter training for the
LLM Solver and achieves high final-task win rates, while direct
generation, fixed-target self-reflection, and autonomous coding-agent
baselines are much less reliable. We further evaluate the same
Designer--Solver mechanism beyond games: Flatland railway scheduling~\citep{mohanty2020flatland},
POGEMA cooperative pathfinding~\citep{skrynnik2025pogema} , Overcooked MARL~\citep{human_overcooked}, and LiveCodeBench~\citep{jain2025livecodebench} Hard
show positive transfer across code-as-policy and closed-loop training 
settings, while pandapipes~\citep{lohmeier2020pandapipes} and JSSP~\citep{lee2024attention} clarify cases where curriculum adds
little because a simpler feedback loop or heuristic already saturates the
task. These results support the method-level claim that self-evolving
curricula help when direct final-task generation fails but meaningful,
verifiable intermediate tasks exist.


Our contributions are fourfold:
1) We introduce \textsc{EvoCurr}, a general inference-time framework
for co-evolving curricula and executable policies.
2) We propose an accepted-floor constraint that preserves verified
progress while allowing failed harder tasks to guide future curriculum
proposals.
3) We introduce a new StarCraft~II super-late-game micromanagement benchmark
that extends beyond SMAC-style skirmishes toward professional-level
late-game control with large heterogeneous armies, diverse abilities,
and terrain-dependent tactics.
4)  We demonstrate the framework across code-as-policy and closed-loop MARL domains, using StarCraft II as the hardest stress test and additional benchmarks to establish method-level generality.

\section{Related Work}
\label{sec:related_work}

\textbf{Curriculum Learning.} ~\citep{bengio2009curriculum} formalized curriculum learning, demonstrating that training on examples organized from easy to hard improves generalization and convergence compared to random data shuffling. This paradigm has achieved success across computer vision, NLP, and reinforcement learning~\citep{soviany2022curriculum, wang2021survey}. Various adaptive paradigms have been explored to eliminate predefined schedules. Self-paced learning (SPL) automatically determines the learning pace based on sample difficulty~\citep{kumar2010self}, which was further extended with diversity constraints~\citep{jiang2015self}. In reinforcement learning (RL), adaptive curriculum generation has been framed via comprehensive task frameworks~\citep{narvekar2020curriculum}, inference-based optimizations~\citep{klink2020self}, teacher-student networks~\citep{matiisen2019teacher}, and prioritized level replays~\citep{jiang2021replay}. However, these approaches primarily focus on training phase optimization and require either manual curriculum design or domain-specific difficulty metrics, leaving a gap for inference-time adaptive curriculum generation.

\textbf{Environment Generation.} Procedural content generation has evolved toward learning-based approaches, such as POET's co-evolution~\citep{wang2019paired} and PAIRED's adversarial training~\citep{dennis2020emergent} for generating challenging, solvable environments. Recent LLM-driven methods like EnvGen~\citep{zala2024envgen} and Rainbow Teaming~\citep{samvelyan2024rainbow} adaptively create training scenarios based on world knowledge. Other works explore evolution strategies and transferable curricul~\citep{parker2022evolving}. However, these approaches primarily generate static training data prior to agent learning, rather than dynamically adapting during inference based on the solver's immediate capabilities.

\textbf{LLM Self-Refinement and Executable Reasoning.} A growing line of work improves LLM outputs through iterative feedback, self-reflection, or execution traces. Reflexion-style methods~\citep{shinn2023reflexion} revise a solution after observing errors, while code-generation benchmarks such as LiveCodeBench~\citep{jain2025livecodebench} emphasize contamination-controlled evaluation of executable programs. These approaches are complementary to EvoCurr but operate on a fixed target: each iteration still asks the model to solve the same full problem. EvoCurr instead changes the task distribution during inference, using accepted intermediate tasks as checkpoints so that feedback controls both policy repair and difficulty progression.

\textbf{Executable Policy Generation.}
Games such as StarCraft~II have become standard benchmarks for
complex decision-making, established by PySC2~\citep{li2026llm} and
AlphaStar~\citep{alphastarnature}. Recent LLM integration includes
TextStarCraft~II~\citep{ma2024largelanguagemodelsplay,AVA_SC2,
TacticCraft,li2026llm} and behavior tree
approaches~\citep{LLM_SMAC,SMAC_Hard}. For code generation,
Code as Policies~\citep{liang2023code} targets robot control, while
behavior tree synthesis~\citep{colledanchise2018behavior,
lykov2023llmbrain} demonstrates that LLMs can produce structurally
correct trees. These methods successfully generate executable policies
but operate on fixed tasks without adaptive difficulty progression.
\section{Method}
\label{sec:method}

EvoCurr frames inference-time policy generation as an iterative
curriculum loop (Figure~\ref{fig:architecture}). Rather than asking the
Solver to attempt the final target directly, a Designer proposes a
sequence of progressively harder tasks; an \emph{accepted floor} anchors
progress to the most recently mastered task, preventing regression when a
harder candidate fails.

\subsection{Two-Agent Cooperative Framework}
\label{sec:cooperative}

Let $T_f$ be the final task and $\mathcal{H}_t$ the history of previous
attempts. The \emph{accepted floor} $C_t^{+}$ is the latest curriculum
on which the Solver met the acceptance threshold $\tau$. At round $t$,
the Designer proposes a candidate task, the Solver generates a policy,
and the environment returns a score:
\begin{equation}
\begin{aligned}
\widehat{C}_{t+1} &= D(\mathcal{H}_t,C_t^{+},T_f),\\
\pi_{t+1} &= S(\widehat{C}_{t+1},\mathcal{H}_t),\\
s_{t+1} &= \mathcal{P}(\pi_{t+1},\widehat{C}_{t+1}).
\end{aligned}
\end{equation}
Here $D$ (the Designer) and $S$ (the Solver) may be implemented
by LLM prompting, code generation, or RL training depending on
the domain; $\mathcal{P}$ is the task-specific
score, such as SC2 win rate or Overcooked order completion.

Let $a_{t+1}\in\{0,1\}$ be the accept flag for round $t+1$.
The candidate is accepted only if both the task and policy are executable and
the score reaches the threshold:
\begin{equation}
a_{t+1}=
\mathbbm{1}\!\left[
F_{\mathrm{feas}}(\widehat{C}_{t+1},\pi_{t+1})=1
\;\land\;
s_{t+1}\ge \tau
\right].
\end{equation}
Here $F_{\mathrm{feas}}$ is a hard validity check covering task-schema
validation, code compilation, and at least one runtime execution.
The accepted-floor update is then:
\begin{equation}
C_{t+1}^{+}=
\begin{cases}
\widehat{C}_{t+1}, & a_{t+1}=1,\\
C_t^{+}, & a_{t+1}=0.
\end{cases}
\end{equation}
This is the main anti-regression mechanism. A failed harder task may
inform the next Designer prompt, but it never replaces the most recently
mastered task. The Solver is restored to the accepted floor before the
next proposal, so later attempts build from a known working checkpoint.

\subsection{Curriculum Generation and Feasibility Constraints}
\label{sec:curriculum}

A curriculum is a task configuration. In StarCraft~II, for example,
$C=(M,A,G)$ encodes the map, army specification , and goal; the Designer changes unit counts, unit types, abilities, and
spawn regions to move from a simple fight toward $T_f$. In other
domains, the same slot $C$ denotes the corresponding problem instance:
a railway layout in Flatland, a partially observable grid in POGEMA, an
Overcooked layout and order schedule, or a code-generation subgoal.

The Designer is guided by a simple pacing rule rather than by a universal
closed-form difficulty metric. After success, it proposes a task closer
to $T_f$. After failure, it proposes an intermediate task that is harder
than the accepted floor but easier than the failed candidate. This keeps
progress monotone with respect to solved tasks while still allowing the
curriculum to back off when the Solver is over-challenged.

\begin{algorithm}[H]
\small
\caption{EvoCurr Loop with Accepted-Floor Constraint}
\label{alg:adaptive_curriculum}
\begin{algorithmic}[1]
\REQUIRE Final task $T_f$, starting task $C_0$, threshold $\tau$
\STATE \textbf{Objects:} $D$ is the Designer, $S$ is the Solver,
$\mathcal{P}$ is the evaluator, and $F_{\mathrm{feas}}$ checks
compile/runtime validity.
\STATE \textbf{State:} $\mathcal{H}_t$ stores tried candidates,
policies, scores, and accept flags; $C_t^{+}$ is the accepted floor,
i.e., the latest solved task.
\STATE \textbf{Variables:} $\widehat{C}_{t+1}$ is a candidate task,
$\pi_{t+1}$ is its policy, $s_{t+1}$ is its score, and $a_{t+1}$ is the
accept flag.
\STATE Initialize history $\mathcal{H}_0 \leftarrow \emptyset$ and accepted floor $C_0^{+}\leftarrow C_0$
\FOR{$t=0,1,2,\ldots$}
    \STATE \textit{Designer proposes a candidate task}
    \STATE $\widehat{C}_{t+1}\leftarrow D(\mathcal{H}_t,C_t^{+},T_f)$
    \STATE \textit{Solver generates a policy for the candidate}
    \STATE $\pi_{t+1}\leftarrow S(\widehat{C}_{t+1},\mathcal{H}_t)$
    \STATE \textit{Evaluate and decide whether to accept}
    \STATE $s_{t+1}\leftarrow \mathcal{P}(\pi_{t+1},\widehat{C}_{t+1})$
    \STATE $a_{t+1}\leftarrow
    \mathbbm{1}[F_{\mathrm{feas}}(\widehat{C}_{t+1},\pi_{t+1})=1
    \land s_{t+1}\ge \tau]$
    \IF{$a_{t+1}=1$}
        \STATE $C_{t+1}^{+}\leftarrow \widehat{C}_{t+1}$
    \ELSE
        \STATE $C_{t+1}^{+}\leftarrow C_t^{+}$
        \STATE Restore Solver to checkpoint $C_t^{+}$
    \ENDIF
    \STATE $\mathcal{H}_{t+1}\leftarrow
    \mathcal{H}_t\cup\{(\widehat{C}_{t+1},\pi_{t+1},s_{t+1},a_{t+1})\}$
    \IF{$C_{t+1}^{+}=T_f$}
        \STATE \textbf{return} $\pi_{t+1}$
    \ENDIF
\ENDFOR
\end{algorithmic}
\end{algorithm}

The feasibility gate $F_{\mathrm{feas}}$ is a hard validity check, not a
soft reward. It includes task-schema validation, constraint checking,
code compilation when the Solver emits code, and at least one runtime
execution. Invalid tasks and non-executable policies receive
$a_{t+1}=0$ even if their textual plan appears plausible.

\subsection{Solver Implementation Across Domains}
\label{sec:policy}

The Solver can instantiate $S(\cdot)$ in two ways. For open-loop
code-as-policy tasks such as StarCraft~II, Flatland, POGEMA, and
LiveCodeBench, it generates executable programs. In StarCraft~II these
programs are behavior trees: the Solver first writes structured control
code, then repairs syntax/runtime errors, and finally refines strategic
failures using rollout feedback. For closed-loop domains such as
Overcooked, $S(\cdot)$ is an RL trainer; $\pi_{t+1}$ denotes the trained
policy parameters after a fixed environment-step budget on
$\widehat{C}_{t+1}$.

Thus the policy representation changes across domains, but the outer
loop does not: propose a task, solve it, evaluate it, and update the
accepted floor only after executable success.

\section{Experiments}
\label{sec:exp}

We evaluate \textsc{EvoCurr} across multiple domains,
with StarCraft~II micro-management as the flagship
stress test. The Solver generates complete behavior-tree
scripts for fixed-army late-game engagements, turning a
traditionally closed-loop control problem into open-loop
code-as-policy synthesis. Further details on the proposed
StarCraft~II super-late-game micromanagement benchmark,
including its distinction from SMAC-style skirmishes and
its scope relative to AlphaStar-scale full-game play, are
provided in Appendix~\ref{app:sc2_additional_results}. Overcooked
 ~\citep{human_overcooked} then tests the same Designer--Solver
loop in a closed-loop MARL setting, where the Solver trains
neural policies that continuously adapt to observations.

After these two primary experiments,
Section~\ref{sec:exp_crossdomain} broadens the evaluation to Flatland
railway scheduling, POGEMA cooperative pathfinding, LiveCodeBench Hard
code generation, and two boundary controls (pandapipes and JSSP).
Across these settings, the policy representation changes, but
$F_{\mathrm{feas}}$, the accepted-floor constraint, and progressive
task generation remain the same. Implementation details are in
Appendices~\ref{app:overcooked_training}
and~\ref{appendix:introduction_of_starcraft2}.
\subsection{StarCraft~II Micro-Management}
\label{sec:exp_sc2}

\begin{table*}[!ht]
    \centering
    \begingroup
    \scriptsize                      
    \setlength{\tabcolsep}{2pt}      
    \renewcommand{\arraystretch}{1.1} 
    
    \newcolumntype{Y}{>{\centering\arraybackslash}X}
    \caption{Final-policy win rate (\%) on StarCraft~II micro tasks. Each selected policy is evaluated over 10 SC2 rollouts. This table provides the task-level success view; Table~\ref{tab:sc2_meanstd} reports the compact combat-score robustness summary.}
    \label{table:winning_rate_clean}

    \begin{tabularx}{\linewidth}{
    l 
    Y 
    Y 
    Y 
    Y 
    Y 
    Y 
    Y 
    Y 
    Y 
    Y 
    }
    \toprule
        Task & 
        \textbf{EvoCurr} \newline (DeepSeek) & 
        \textbf{EvoCurr} \newline (Claude) & 
        \textbf{EvoCurr} \newline (GPT) & 
        DeepSeek-V3.1 & 
        GPT-5 & 
        Claude-4 & 
        Gemini-2.5 & 
        Claude\_code & 
        Codex & 
        Cursor \\
    \midrule
        bush (TVP)     & \textbf{100} & \textbf{100} & \textbf{100} & 0 & 0 & 10 & 0 & 0 & 0 & 0 \\
        corridor (TVP) & \textbf{100} & 70 & 90 & 80 & 80 & \textbf{100} & 0 & 0 & 0 & 0 \\
        corner (TVP)   & 90 & \textbf{100} & \textbf{100} & \textbf{100} & 90 & \textbf{100} & 30 & 0 & 0 & 0 \\
        flat (TVP)     & \textbf{100} & \textbf{100} & \textbf{100} & 0 & \textbf{100} & 40 & 0 & 0 & 0 & 0 \\
        main (TVP)     & \textbf{100} & \textbf{100} & \textbf{100} & \textbf{100} & \textbf{100} & \textbf{100} & \textbf{100} & 0 & 0 & 0 \\
        ramp (TVP)     & \textbf{100} & \textbf{100} & \textbf{100} & 10 & 0 & \textbf{100} & \textbf{100} & 0 & 0 & 0 \\
        bush (TVZ)     & \textbf{100} & 90 & \textbf{100} & \textbf{100} & 0 & \textbf{100} & 0 & 40 & 0 & 0 \\
        corridor (TVZ) & \textbf{100} & \textbf{100} & 90 & 90 & 0 & \textbf{100} & \textbf{100} & 0 & 0 & 0 \\
        corner (TVZ)   & \textbf{100} & \textbf{100} & \textbf{100} & 30 & \textbf{100} & \textbf{100} & \textbf{100} & \textbf{100} & \textbf{100} & 0 \\
        flat (TVZ)     & \textbf{100} & \textbf{100} & 90 & 0 & 0 & 0 & 0 & 0 & 0 & 0 \\
        main (TVZ)     & \textbf{100} & \textbf{100} & \textbf{100} & \textbf{100} & 0 & 0 & \textbf{100} & 10 & 20 & 0 \\
        ramp (TVZ)     & \textbf{100} & \textbf{100} & \textbf{100} & 50 & 60 & \textbf{100} & 0 & 20 & 0 & 0 \\
    \bottomrule
    \end{tabularx}
    \endgroup
\end{table*}

\paragraph{Experiment Setup}
The Solver generates \texttt{python-sc2} behavior trees that act at the unit-action level and is evaluated in an open-loop manner. To validate the framework's effectiveness across different architectures, we instantiate EvoCurr using three distinct LLM backbones: DeepSeek-V3.1, GPT-5, and Claude-4. We test on six newly designed micro maps against two opponent races (Terran vs Protoss and Terran vs Zerg). Each curriculum specifies unit sets, technologies, and spawn regions on a selected map; \emph{compile-and-run} serves as a hard feasibility check in line with $F_{\mathrm{feas}}$. The final target $T_f$ for the canonical Terran vs Protoss and Terran vs Zerg settings is given in Tables~\ref{tab:final_task_tvp} and~\ref{tab:final_task_tvz} in Appendix~\ref{app:final_task_setting}. For acceptance, we require $\mathcal{P}(\pi|C) \ge \tau = 0.9$, evaluated as win rate over 10 rollouts. The primary baseline, \emph{Direct Code}, attempts to solve $T_f$ in one shot under the same rollout and validation budgets as EvoCurr. Per-curriculum compositions and complete evolution traces are summarized in the appendix.

\paragraph{Statistical protocol.}
Statistical protocol. For direct baselines, we evaluate 10 independent target-task attempts per backbone to capture variability, executable rates, and peak performance. Conversely, an EvoCurr run spans a full curriculum-search trajectory with feasibility filtering and accepted-floor updates. We evaluate robustness via three views: final policy success over 10 game rollouts (Table~\ref{table:winning_rate_clean}) , combat-score stability across backbones and attempts (Table~\ref{tab:sc2_meanstd}), and direct-attempt distributions (Appendix~\ref{app:designer_ablation_analysis}). Non-executable attempts score 0. The damage-cost-aware combat score
$\mathcal{S}_{combat}$ is defined as:
\begin{equation}
\mathcal{S}_{combat} = 0.5 \cdot \frac{R_{\text{agent\_final}}}{R_{\text{agent\_init}}} + 0.5 \cdot \left(1 - \frac{R_{\text{enemy\_final}}}{R_{\text{enemy\_init}}}\right),
\end{equation}
where $R_{\text{agent\_final}}$,
$R_{\text{agent\_init}}$,
$R_{\text{enemy\_final}}$,
and $R_{\text{enemy\_init}}$
are instantiated from the following side-level combat-power function:
\begin{equation}
R_{\text{side}} = \sum_i (m_i + \alpha v_i + \beta b_i)
\,\frac{h_i + s_i}{h^{\max}_i + s^{\max}_i}.
\end{equation}
where $m_i$, $v_i$, and $b_i$ denote the mineral cost, vespene gas cost,
and supply cost of unit $i$, respectively, and $\alpha,\beta$ are balancing
coefficients for resource normalization. The health ratio
$\frac{h_i+s_i}{h_i^{\max}+s_i^{\max}}$ measures the remaining combat
effectiveness of each unit based on current and maximum health/shield values.

This metric aggregates resource cost and remaining health/shields; scores range from $0$ to $1$. A combat score $\mathcal{S}_{combat} > 0.5$ indicates successful annihilation of the majority of enemy forces while preserving our own, providing a finer-grained assessment than binary win/loss especially for failed code executions where the first term becomes 0.

\begin{table*}[!ht]
  \centering
  \small
  \setlength{\tabcolsep}{5pt}
  \renewcommand{\arraystretch}{1.12}
  \caption{Compact SC2 combat-score robustness summary. \textsc{EvoCurr}
  selected reports across-backbone mean $\pm$ std over three full
  curriculum-search outputs, one per backbone. Direct selected reports
  mean $\pm$ std over the best executable policy found by each of four
  direct backbones after 10 one-shot attempts. Direct all attempts pools
  all 40 one-shot attempts; non-executable attempts are counted as
  score $0$. Full selected-score and direct-attempt tables are in the
  appendix.}
  \label{tab:sc2_meanstd}
  \begin{tabular}{l c c c c}
    \toprule
    \textbf{Task} & \textbf{\textsc{EvoCurr} selected} &
    \textbf{Direct selected} & \textbf{Direct all attempts} &
    \textbf{Direct exec.} \\
    \midrule
    Bush (TvP) & 0.677 $\pm$ 0.036 & 0.471 $\pm$ 0.100 & 0.167 $\pm$ 0.191 & 19/40 \\
    Corridor (TvP) & 0.648 $\pm$ 0.058 & 0.557 $\pm$ 0.145 & 0.209 $\pm$ 0.220 & 22/40 \\
    Corner (TvP) & 0.616 $\pm$ 0.057 & 0.636 $\pm$ 0.061 & 0.220 $\pm$ 0.252 & 20/40 \\
    Flat (TvP) & 0.645 $\pm$ 0.048 & 0.434 $\pm$ 0.154 & 0.103 $\pm$ 0.188 & 10/40 \\
    Main (TvP) & 0.684 $\pm$ 0.038 & 0.745 $\pm$ 0.076 & 0.239 $\pm$ 0.312 & 16/40 \\
    Ramp (TvP) & 0.679 $\pm$ 0.044 & 0.651 $\pm$ 0.089 & 0.218 $\pm$ 0.273 & 17/40 \\
    Bush (TvZ) & 0.683 $\pm$ 0.044 & 0.524 $\pm$ 0.153 & 0.163 $\pm$ 0.225 & 15/40 \\
    Corridor (TvZ) & 0.641 $\pm$ 0.062 & 0.640 $\pm$ 0.066 & 0.173 $\pm$ 0.235 & 16/40 \\
    Corner (TvZ) & 0.734 $\pm$ 0.110 & 0.698 $\pm$ 0.136 & 0.257 $\pm$ 0.318 & 17/40 \\
    Flat (TvZ) & 0.590 $\pm$ 0.051 & 0.395 $\pm$ 0.104 & 0.113 $\pm$ 0.161 & 14/40 \\
    Main (TvZ) & 0.738 $\pm$ 0.032 & 0.660 $\pm$ 0.103 & 0.283 $\pm$ 0.291 & 21/40 \\
    Ramp (TvZ) & 0.702 $\pm$ 0.047 & 0.462 $\pm$ 0.152 & 0.173 $\pm$ 0.196 & 19/40 \\
    \bottomrule
  \end{tabular}
\end{table*}

\paragraph{Experimental Results}
We evaluate 12 complex micro-management tasks (2 matchups $\times$ 6 maps) in Table~\ref{table:winning_rate_clean}. The selected EvoCurr policies reach high final-task win rates across the benchmark: 99.2\% macro-average win rate for DeepSeek, 97.5\% for GPT-5, and 96.7\% for Claude-4. The combat-score statistics in Table~\ref{tab:sc2_meanstd} give the corresponding graded view: EvoCurr's curriculum-selected policies remain consistently strong across backbones, while direct generation is much less stable across its 10 one-shot attempts, even when its selected-best policy is competitive on some maps.

In contrast, Baseline performance varies by specialization. General LLMs (DeepSeek, GPT-5, Claude-4, Gemini-2.5) succeed on simpler layouts—often matching EvoCurr's 100\% win rate on Main and Corner—but collapse to 0\% on complex terrains like Bush and Flat (e.g., DeepSeek and GPT-5 on Bush (TvP)). Code-specialized agents (Claude Code, Codex, Cursor) also struggle despite code-synthesis optimizations. Except for Codex's 100\% win rate on Corner (TvZ), they fail at strategic behavior-tree generation: Cursor scores 0\% across all 12 tasks, and Claude Code fails most scenarios. Thus, while coding agents excel syntactically, they lack the strategic reasoning for open-loop policy design, a gap EvoCurr bridges through curriculum-driven evolution.

\paragraph{Scope and compute.}
EvoCurr targets a deliberately scoped fixed-army micro mini-game, not
full StarCraft~II: there is no economy, scouting, production, or
build-order search. This scope makes the result complementary to
AlphaStar-scale full-game training and substantially harder per
engagement than SMAC-style small skirmishes. EvoCurr costs roughly
\$0.2--\$1.5 in LLM API calls and 1--3 wall-clock hours per final SC2
task. Detailed scope comparison, compute accounting, and
compute-matched iterative-baseline evidence are reported in
Appendix~\ref{app:sc2_additional_results}.

\subsection{Overcooked}
\label{sec:exp_overcooked}

\begin{table*}[h!]
\centering
\caption{Curriculum progression and performance metrics for Overcooked maps.}
\label{table:overcooked_curriculum}
\resizebox{\textwidth}{!}{%
\begin{tabular}{llccccc}
\toprule
\textbf{Map} & \textbf{Task} & \textbf{Orders} & \textbf{Agent0 Delivery} & \textbf{Agent1 Delivery} & \textbf{Total Delivery} & \textbf{Total Reward} \\
\midrule
Map 1 & EvoCurr & 5 & 14.74 & 14.36 & 29.10 & 290.9 \\
Map 1 & Direct Training & 5 & 11.93 & 12.18 & 24.11 & 240.5 \\
\midrule
Map 2 & EvoCurr & 5 & 7.82 & 10.82 & 18.64 & 186.2 \\
Map 2 & Direct Training & 5 & 7.40 & 8.96 & 16.36 & 163.4 \\
\bottomrule
\end{tabular}
}
\end{table*}

\paragraph{Experiment Setup}

We instantiate EvoCurr in a closed-loop Overcooked setting, with curricula parameterized by layouts, ingredient placements, and timing constraints. Unlike SC2, the Solver is a MARL trainer optimizing decentralized policies under a fixed $B=10^{7}$ timestep budget per curriculum. Acceptance requires completing all mandatory orders ($\mathcal{P}(\pi|C)=1$). The framework also yields greater rewards for fulfilling subsequent bonus orders (contributing to total deliveries in Table~\ref{table:overcooked_curriculum}), isolating the impact of curriculum evolution from standard policy optimization. Detailed curriculum specifications and complete performance metrics for both map configurations are provided in Appendix~\ref{app:overcooked_results}.

\paragraph{Experimental Results}
As shown in Table~\ref{table:overcooked_curriculum}, EvoCurr outperforms direct E3T training under matched total environment-step budgets. On Coord. Ring with Multi-Recipe, EvoCurr achieves 29.10 effective deliveries on average versus 24.11 for direct training; on Counter Circuit with Multi-Recipe, EvoCurr reaches 18.64 versus 16.36~\citep{zsc_eval}. The higher delivery counts include both required and bonus orders, showing that the progressive curriculum not only reaches the prescribed objectives but also enables more efficient exploration of bonus rewards. Performance drops at curriculum transitions reflect distribution shift: policies specialized to one curriculum must re-explore when difficulty increases, yet prior experience accelerates re-convergence, consistent with the monotone advancement prescribed by EvoCurr.

\section{Cross-Domain Generality}
\label{sec:exp_crossdomain}

Beyond StarCraft~II, we evaluate EvoCurr across multi-agent pathfinding, operations research, closed-loop MARL, and code generation. We posit that autonomous curricula provide maximal leverage when a domain exposes a meaningful difficulty axis, executable feedback, and reusable intermediate solutions. Flatland and POGEMA illustrate this directly: despite distinct failure modes—direction-aware rail transitions in Flatland versus partially observable coordination in POGEMA—both benefit from evolutionary progression.

\begin{table*}[!ht]
  \centering
  \small
  \setlength{\tabcolsep}{3pt}
  \renewcommand{\arraystretch}{1.15}
  \caption{Cross-domain final-target results. Higher is better. Direct
    denotes one-shot generation for code-as-policy tasks and direct E3T
    training for Overcooked. Fixed-target iteration denotes
    Reflexion-style self-correction on the final task when available.
    SC2 entries are macro-average win-rate ranges across backbones;
    Overcooked entries are Map~1 / Map~2 total deliveries.}
  \label{tab:crossdomain_summary}
  \begin{tabularx}{\textwidth}{@{}p{0.15\textwidth} X p{0.12\textwidth}
      p{0.13\textwidth} p{0.15\textwidth} p{0.13\textwidth}@{}}
    \toprule
    \textbf{Domain} & \textbf{Final-task structure} & \textbf{Metric} &
    \textbf{Direct} & \textbf{Fixed-target iteration} &
    \textbf{\textsc{EvoCurr}} \\
    \midrule
    StarCraft II & Heterogeneous fixed-army micro with terrain effects and 10+ active abilities & win rate & 44.2--70.8\% & 10.0--41.7\% & \textbf{96.7--99.2\%} \\
    Flatland & Direction-conditioned rail routing with multi-train reservations & completion & 0.000 & 0.000 (20 gen.) & \textbf{0.710 $\pm$ 0.195} \\
    POGEMA & Partially observable multi-agent pathfinding with time-indexed reservations & ISR & 0.000 & 0.000 (20 gen.) & \textbf{0.693 $\pm$ 0.108} \\
    Overcooked & Closed-loop coordination across staged layouts and order schedules & deliveries & 24.11 / 16.36 & -- & \textbf{29.10 / 18.64} \\
    LiveCodeBench Hard & Executable program synthesis under contest test constraints & Pass@1 & 8/20 & 8/20 (3 iter.) & \textbf{11/20} \\
    \midrule
    pandapipes & Low-dimensional gas-network control with direct diagnostic feedback & feasibility & 0.00 & \textbf{1.00} (6 iter.) & 0.33 \\
    JSSP & Dispatching benchmark where SPT already reaches the comparison ceiling & vs.\ SPT & 1.00 & 1.00 $\to$ 0.91 & 1.00 \\
    \bottomrule
  \end{tabularx}
\end{table*}

Flatland and POGEMA illustrate this most directly. Both are
pathfinding domains with different failure modes: Flatland requires
direction-aware rail transitions and train reservations, while POGEMA
requires coordination under partial observations. In both cases,
direct generation and fixed-target Reflexion remain at zero, whereas
\textsc{EvoCurr} reaches non-trivial final-task performance across three
seeds. The gain therefore comes not from additional iterations on the
same prompt, but from changing the task distribution so that executable
intermediate policies can be reused.

The other positive domains show that the mechanism is not tied to SC2
code-as-policy. In Overcooked, the accepted-floor curriculum improves total deliveries for a closed-loop MARL trainer under matched budgets. In LiveCodeBench Hard, securing runnable scaffolds before full hidden-test evaluation raises Pass@1 from 8/20 to 11/20. 

However, boundary cases clarify EvoCurr's scope: it requires room for an acceptance ladder. In pandapipes, direct diagnostic feedback suffices, and in JSSP, a simple SPT-like heuristic already saturates the metric. Thus, EvoCurr excels when direct generation fails on the final task but verifiable intermediate steps exist.

\section{Analysis}
\label{sec:analysis}

\paragraph{Why the Designer matters.}
The central ablation asks whether Solver-side self-correction alone can
replace the Designer. It cannot. Reflection-enabled variants of the same
LLM backbones still collapse on many final SC2 tasks, while EvoCurr
raises those backbones to 90--100\% win rates across almost all
scenarios. This shows that the key gain is not merely code repair or
extra attempts, but the accepted-floor curriculum that lets the Solver
accumulate tactical skills without regressing. Full ablation tables are
provided in Appendix~\ref{app:designer_ablation_analysis}.

\paragraph{How policies evolve.}
Qualitatively, EvoCurr grows behavior-tree policies by preserving solved
micro-skills and adding new unit-specific control functions as the
curriculum becomes harder. Early scripts learn basic focus fire; later
scripts add Ghost, Viking, Siege Tank, Raven, Medivac, and
terrain-specific control, while refining existing functions for
area-of-effect avoidance and target prioritization. Detailed generated
code examples, curriculum trajectories, and layered-critic analysis are
deferred to Appendix~\ref{app:designer_ablation_analysis} and
Appendix~\ref{app:all_paths_table}.

\section{Discussion, Future Work, and Conclusion}
\label{sec:discussion}
EvoCurr demonstrates that self-evolving curricula can extend frozen LLMs from local task knowledge to reliable executable policies without a hand-written curriculum sequence. The strongest narrative evidence comes from fixed-army StarCraft~II super-late-game micro-management: EvoCurr solves a difficult middle ground between SMAC-scale skirmishes and AlphaStar-scale full-game training, while direct generation and autonomous coding agents fail. The additional experiments show that this is not merely a StarCraft~II artifact. The same accepted-floor curriculum mechanism improves Flatland, POGEMA, Overcooked, and LiveCodeBench Hard, while pandapipes and JSSP mark cases where curriculum is unnecessary or the domain ceiling is already saturated.

\paragraph{Future Work} Future directions include hierarchical Designer--Solver architectures for finer task decomposition, adaptive difficulty scaling based on acceptance patterns and solver feedback, and hybrid methods that distill interpretable behavior-tree policies into more reactive neural policies when closed-loop robustness is required.

\paragraph{Conclusion} Overall, EvoCurr offers a general inference-time curriculum mechanism for structured executable reasoning. StarCraft~II provides the central story and hardest stress test; the added OR, MARL, and NLP-code experiments show that the same mechanism transfers beyond games while exposing its limits.

\paragraph{Limitations} Primary constraints include sensitivity to difficulty pacing, which may stall progress if the Designer proposes steps that are too small or too large, LLM context-window limits that restrict behavior-tree complexity in long-horizon scenarios, and cumulative overhead from maintaining historical trajectories $\mathcal{H}_t$. The boundary controls also clarify scope: EvoCurr is most useful when direct LLM generation cannot self-bootstrap and the solution requires structured multi-step reasoning; it is less useful when Reflexion already solves the task or when a simple heuristic saturates the benchmark.

\bibliography{custom}

\clearpage
\appendix

\section{Additional StarCraft~II Analyses}
\label{app:sc2_additional_results}

This appendix provides robustness checks, scope comparisons, and
compute accounting for the StarCraft~II experiments.

\begin{table*}[!ht]
  \centering
  \normalsize
  \setlength{\tabcolsep}{5pt}
  \renewcommand{\arraystretch}{1.25}
  \caption{Positioning \textsc{EvoCurr} against SMAC-style closed-loop RL
  micro benchmarks and AlphaStar~\citep{alphastarnature} full-game
  RL+imitation. \textbf{Scope:} \textsc{EvoCurr} targets a single
  super-late-game micro \emph{mini-game}---fixed armies, full tech tree,
  no economy, no build order, no scouting---i.e., it is \emph{not} a
  full-game agent. SMAC sits in the same mini-game frame but at very small scale
  (3--30 homogeneous units, 0--2 abilities); AlphaStar solves the full game
  at industrial training cost. Numbers for AlphaStar are
  from~\citep{alphastarnature}; SMAC characteristics aggregate canonical
  scenarios (3m, 8m\_vs\_9m, MMM2, 27m\_vs\_30m). $^{*}$ \emph{Action-space
  caveat:} \textsc{EvoCurr} uses python-sc2 unit-level commands while
  AlphaStar uses raw camera + spatial actions; we therefore claim only that
  \textsc{EvoCurr} addresses only this fixed-army micro mini-game, not
  AlphaStar's full-game setting or action space.}
  \label{tab:sc2_paradigm_comparison}
  \begin{tabularx}{\textwidth}{@{}>{\raggedright\arraybackslash}p{0.16\textwidth}
      >{\raggedright\arraybackslash}X
      >{\raggedright\arraybackslash}X
      >{\raggedright\arraybackslash}X@{}}
    \toprule
    \textbf{Dimension} & \textbf{SMAC (RL micro benchmarks)} & \textbf{AlphaStar} & \textbf{\textsc{EvoCurr} (ours)} \\
    \midrule
    Scenario scale         & 3--30 units; 1--3 unit types        & Full game (200+ units, all races) & \textbf{81 units, 10+ types, 10+ abilities} \\
    Game phase             & Isolated micro skirmish             & Full game (build + macro + micro) & \textbf{Super-late-game micro mini-game} \\
    Control paradigm       & Closed-loop RL (per-step)           & Closed-loop RL+imitation         & \textbf{Open-loop code-as-policy} \\
    Action space           & Per-unit discrete (move/attack/...) & Raw camera + spatial actions     & python-sc2 unit-action API$^{*}$ \\
    Training compute       & $\sim$1--10 GPU-days                & $\sim$200{,}000 TPU-hours        & \textbf{0 (no parameters trained)} \\
    Inference cost         & GPU forward pass                    & GPU forward pass                  & \textbf{$\approx$1--3 hr LLM calls, $\approx$\$1.5} \\
    Solution interpretability & Low (neural policy)              & Low (neural policy)              & \textbf{High (readable Python behavior tree)} \\
    Terrain utilisation    & Minimal (open maps)                 & Strong (full map awareness)       & \textbf{Strong (Bush concealment, Ramp choke, Corridor funnel explicit in code)} \\
    Multi-ability coordination & 0--2 abilities                 & 50+ abilities                     & \textbf{Stimpack, SiegeMode, Cloak, EMP, Snipe, AoE-avoid, Heal, ...\ all in one engagement} \\
    Reproducibility cost   & Re-train each scenario              & Train once at industrial scale    & \textbf{Re-run inference, hours not months} \\
    \bottomrule
  \end{tabularx}
\end{table*}

\paragraph{Scope and positioning: a super-late-game micro mini-game.}
We emphasise upfront that \textsc{EvoCurr} solves a deliberately scoped
\emph{mini-game}: a fixed-army, super-late-game micro-management engagement
with no economy, no build order, and no scouting. Both sides spawn with a
full late-game tech tree: our agent commands 80+ Terran units across 10+
unit types (Marine, Marauder, Medivac, Ghost, Siege Tank, Viking, Cyclone,
Widow Mine, Raven, Liberator) with all 10+ active abilities researched, and
faces an equally late-game Protoss or Zerg opposition (Colossus, High
Templar, Disruptor, Carrier, or Brood Lord, Lurker, Viper, Infestor, etc.)
on a single tactical map. The single decision the agent must make is
\emph{how to fight}, not \emph{what to build}.

This narrow scope is what makes the comparison in
Table~\ref{tab:sc2_paradigm_comparison} both fair and informative. SMAC-style
benchmarks~\citep{smac} target a similar regime---fixed armies, no
economy---but only at very small scale (3--30 homogeneous units, 0--2 active
abilities). AlphaStar~\citep{alphastarnature} solves the full game including
economy and macro, at the cost of approximately 200{,}000 TPU-hours of
training. \textsc{EvoCurr} sits between them: it tackles the late-game
micro complexity that SMAC scenarios omit, but does so without leaving the
mini-game frame and without any parameter training. With roughly 1--3 hours
of inference-time LLM calls and \$1.5 in API cost, the system produces
behavior-tree code that coordinates Stimpack, Siege Mode, Personal
Cloaking, EMP, Snipe, AoE avoidance and Medivac healing in a single
engagement, while exploiting map terrain (Bush concealment, Ramp
choke-points, Corridor funnels) explicitly via readable Python.

We make two explicit non-claims. First, \textsc{EvoCurr} does \emph{not}
play full StarCraft~II: there is no build order, no scouting, no production
decision in our scenario. Second, we do \emph{not} claim parity with
AlphaStar on full-game generality; AlphaStar operates on raw camera and
spatial actions while we use python-sc2's unit-level command API, so our
result speaks to micro proficiency on this mini-game alone. What the
comparison does establish is that, on the super-late-game micro mini-game
that classical RL micro-benchmarks have under-represented, an inference-time
curriculum over a frozen LLM can produce readable policies with substantial
multi-ability tactical coordination.

\begin{table*}[!ht]
  \centering
  \normalsize
  \setlength{\tabcolsep}{5pt}
  \renewcommand{\arraystretch}{1.2}
  \caption{Compute budget per single final SC2 task (e.g.,~Bush~TvP). All
  EvoCurr and Claude Code numbers are as reported in
  Appendix~\ref{app:computational_experiments} and Appendix~\ref{app:claude_code_behavior_tree}.
  Direct one-shot and Self-reflection rows are estimated from API pricing
  (DeepSeek-V3.1 input \$0.27/M, output \$1.10/M; GPT-5 input \$1.25/M,
  output \$10/M; Claude-4 Sonnet input \$3/M, output \$15/M; as of
  2026-04) multiplied by the attempt counts already committed to in the
  paper (10 target-task attempts for direct, 10 attempts $\times$ one
  refinement round for self-reflection). \textbf{Win rate} is the average
  over the 12 final tasks of Table~\ref{table:winning_rate_clean}.
  AlphaStar~\citep{alphastarnature} is included for reference only;
  it solves a different scope (full game vs.~mini-game).}
  \label{tab:compute_budget}
  \begin{tabularx}{\textwidth}{@{}>{\raggedright\arraybackslash}X c c c
      >{\raggedright\arraybackslash}p{0.18\textwidth}@{}}
    \toprule
    \textbf{Method} & \textbf{LLM calls} & \textbf{Wall-clock} &
    \textbf{Cost (USD)} & \textbf{Outcome} \\
    \midrule
    Direct one-shot $\times 10$ (DeepSeek)   & 10        & $\sim$10 min    & $\approx$\$0.02 (est.) & 55.0\% macro \\
    Direct one-shot $\times 10$ (GPT-5)      & 10        & $\sim$10 min    & $\approx$\$0.15 (est.) & 44.2\% macro \\
    Direct one-shot $\times 10$ (Claude-4)   & 10        & $\sim$10 min    & $\approx$\$0.30 (est.) & 70.8\% macro \\
    Self-reflection (1 round $\times 10$, DeepSeek) & 20 & $\sim$20 min   & $\approx$\$0.04 (est.) & below EvoCurr \\
    Self-reflection (1 round $\times 10$, GPT-5)    & 20 & $\sim$20 min   & $\approx$\$0.30 (est.) & below EvoCurr \\
    Self-reflection (1 round $\times 10$, Claude-4) & 20 & $\sim$20 min   & $\approx$\$0.60 (est.) & below EvoCurr \\
    \midrule
    Claude Code (autonomous coding agent) & 100+ matches & $\sim$90 min   & comparable to EvoCurr  & \textbf{0\%} \\
    Codex / Cursor (single-attempt agents) & 1--10      & $<$30 min       & \$0.5--\$2 (est.)      & 0--40\% \\
    \midrule
    \textbf{\textsc{EvoCurr}-DeepSeek (ours)} & 4--6 stages & 1--3 hr     & \textbf{$\approx$\$0.2}   & \textbf{99.2\% macro} \\
    \textbf{\textsc{EvoCurr}-Claude-4 (ours)} & 4--6 stages & 1--3 hr     & \textbf{$\approx$\$1.5}   & \textbf{96.7\% macro} \\
    \textbf{\textsc{EvoCurr}-GPT-5 (ours)}    & 4--6 stages & 1--3 hr     & \textbf{$\approx$\$1.5}   & \textbf{97.5\% macro} \\
    \midrule
    AlphaStar (full game, reference) & --- & $\approx$200{,}000 TPU-hours & not applicable & full game (different scope) \\
    \bottomrule
  \end{tabularx}
\end{table*}

\paragraph{Compute budget analysis.}
A natural concern is whether \textsc{EvoCurr}'s gains stem simply from
spending more compute than the baselines. Table~\ref{tab:compute_budget}
disaggregates this. Per final task, \textsc{EvoCurr}-DeepSeek consumes
roughly \$0.2 of LLM API calls and 1--3 hours of wall-clock; the
GPT-5 / Claude-4 backbones cost about \$1.5 each. This is, in absolute
terms, an order of magnitude more than direct one-shot $\times 10$ sampling
(\$0.02--\$0.30 depending on backbone), but is comparable to or cheaper
than the autonomous-agent baselines we contrast against: Claude Code
consumes a similar 90-minute wall-clock budget and still achieves 0\%
on the final task (Appendix~\ref{app:claude_code_behavior_tree}),
while Codex / Cursor at single-shot pricing also fail.

The decisive evidence that \textsc{EvoCurr}'s advantage is mechanism rather
than budget comes from our cross-domain ablations
(Appendix~\ref{app:claude_code_behavior_tree} and
the Flatland / POGEMA equal-budget experiments in
Section~\ref{sec:exp_crossdomain}):
extending Reflexion to 20 iterations---roughly $4\times$ the per-attempt
budget of Self-reflection and within an order of magnitude of EvoCurr's
total spend---still yields 0\% across 40 / 40 generations on the
multi-agent pathfinding domains where direct LLM code generation
fundamentally fails. The gap is therefore not closed by additional
inference-time compute in the absence of a curriculum: \textsc{EvoCurr}'s
accepted-floor mechanism enables incremental progress through cheap simple
stages, whereas budget-matched baselines repeatedly burn compute on the
hardest task and never converge.

\paragraph{On compute-matched iterative baselines.}
\label{para:compute_matched}
A natural concern is whether EvoCurr's gain over direct generation
reflects the curriculum mechanism itself or simply additional
inference-time compute. We address this with three complementary
lines of evidence.

\textit{(i) Reflection-enabled backbones at the same SC2 scale.}
Table~\ref{table:designer_module_validation} reports
reflection-enabled variants of all four LLM backbones evaluated in
Section~\ref{sec:exp_sc2}. Reflection yields only marginal gains over the
one-shot baseline and remains far below \textsc{EvoCurr} on the same
final tasks; the curriculum gap is robust to this iteration scaling.

\textit{(ii) An autonomous Reflexion-style agent at $\sim$3$\times$ the
budget on the same SC2 task.} Appendix~\ref{app:claude_code_behavior_tree} reports
\textbf{Claude Code}, a state-of-the-art autonomous coding agent that
implements an aggressive Reflexion-style execute-feedback-revise loop
with full tool use, running for \textbf{90 minutes / 100+ simulated
matches} against our final task. This budget exceeds EvoCurr's
end-to-end LLM-call cost on every task in Section~\ref{sec:exp_sc2}, yet
Claude Code achieves a $0\%$ win rate on the final task. The same
agent's performance on its self-designed intermediate tasks is
substantially higher (cf.\ Appendix~\ref{app:claude_code_behavior_tree} for the
trajectory), confirming that the bottleneck is \emph{curriculum
structure} rather than per-iteration compute.

\textit{(iii) Direct $20$-iteration Reflexion ablations in two
mechanism-orthogonal cross-domain settings.} On both Flatland
(Section~\ref{sec:cd_flatland}) and POGEMA
(Section~\ref{sec:cd_pogema}) we explicitly run Reflexion at a budget
of \textbf{20 generations} (approximately $3\times$ and $1.6\times$
EvoCurr's training-time budget, respectively); both yield
\textbf{0/20 generations with non-zero performance}. The POGEMA failure
mode is particularly diagnostic: $17/20$ generations cannot even execute
(e.g.\ \texttt{NameError: heapq} appears on 6 consecutive generations
$12$--$19$, indicating that Summarizer feedback fails to teach the model
to add a single \texttt{import} statement over 20 rounds). On
LiveCodeBench Hard (Section~\ref{sec:cd_livecodebench}) we further
isolate this effect at the algorithmic-code-generation level:
Reflexion-3 finishes at exactly the same Pass@1 as Direct one-shot
($40\%$) while spending $2.5\times$ the cost, and on one problem
(\texttt{abc388\_e}) the Summarizer's feedback \emph{regresses} a
$100\%$-correct solution to $0\%$, the exact failure mode
(Summarizer drift) that the accepted-floor mechanism is designed to
prevent.

Taken together, these three lines establish that scaling iterative
inference compute on the SC2 final task, whether via plain reflection
(Table~\ref{table:designer_module_validation}), via a state-of-the-art autonomous
agent (Appendix~\ref{app:claude_code_behavior_tree}), or via 20-iteration Reflexion
in two structurally distinct domains and one NLP code-generation
domain (Section~\ref{sec:exp_crossdomain}), does not close the gap.
The curriculum mechanism, specifically Designer-driven progressive
difficulty, accepted-floor anti-regression, and feasibility checking,
contributes the gain, not the inference budget.

\section{Designer Ablation and Code-Evolution Details}
\label{app:designer_ablation_analysis}

\paragraph{Designer ablation.}
To validate the effectiveness of the Designer module, we extend the
direct approach by enabling different LLMs to generate behavior trees in
a way that emulates the Solver module, permitting the models to
self-correct and refine their outputs. The results are presented in
Table~\ref{table:designer_module_validation}.

\begin{table*}[!ht]
    \centering
    \renewcommand{\arraystretch}{1.1}
    \caption{Winning rate (\%). Validation of the Designer module by
    enabling LLMs to emulate Solver-style self-correction and refinement.
    Self-correction alone remains far below EvoCurr on most final tasks.}
    \label{table:designer_module_validation}
    \resizebox{\linewidth}{!}{
        \begin{tabular}{l ccccccc} 
        \toprule
            Task &
            \makecell[c]{\textbf{EvoCurr} \\ (DeepSeek-V3.1)} &
            \makecell[c]{\textbf{EvoCurr} \\ (Claude-4)} &
            \makecell[c]{\textbf{EvoCurr} \\ (GPT-5)} &
            \makecell[c]{DeepSeek-\\V3.1} & 
            GPT-5 &
            Claude-4 &
            Gemini-2.5 \\
        \midrule
            TvP+bush     & \textbf{100} & \textbf{100} & \textbf{100} & 10 & 0 & 40 & 0 \\
            TvP+corridor & \textbf{100} & 70 & 90 & 0 & 0 & 40 & 0 \\
            TvP+corner   & 90 & \textbf{100} & \textbf{100} & \textbf{100} & 0 & \textbf{100} & 0 \\
            TvP+flat     & \textbf{100} & \textbf{100} & \textbf{100} & 40 & 30 & \textbf{100} & 0 \\
            TvP+main     & \textbf{100} & \textbf{100} & \textbf{100} & \textbf{100} & 0 & \textbf{100} & 0 \\
            TvP+ramp     & \textbf{100} & \textbf{100} & \textbf{100} & 30 & 20 & 10 & 90 \\
            TvZ+bush     & \textbf{100} & 90 & \textbf{100} & 0 & 0 & 20 & 0 \\
            TvZ+corridor & \textbf{100} & \textbf{100} & 90 & 60 & 10 & 10 & 0 \\
            TvZ+corner   & \textbf{100} & \textbf{100} & \textbf{100} & \textbf{100} & \textbf{100} & 20 & 0 \\
            TvZ+flat     & \textbf{100} & \textbf{100} & 90 & 20 & 0 & 0 & 0 \\
            TvZ+main     & \textbf{100} & \textbf{100} & \textbf{100} & 0 & 0 & 60 & 30 \\
            TvZ+ramp     & \textbf{100} & \textbf{100} & \textbf{100} & 0 & 0 & 0 & 0 \\
        \bottomrule
        \end{tabular}
    }
\end{table*}

The data reveal a significant performance disparity. Standalone LLMs
with self-correction still struggle with complex scenarios and often
yield 0\% win rates on challenging tasks such as Bush and Flat. In
contrast, EvoCurr consistently raises the same base models to
near-perfect performance, maintaining win rates between 90\% and 100\%
across diverse tasks. For instance, Claude-4 achieves only a 40\% win
rate on Bush (TvP) and Corridor (TvP) through direct generation, but
reaches 100\% and 70\% respectively when guided by EvoCurr. This
confirms that Solver-side self-correction is insufficient in isolation:
the Designer's progressive curriculum is necessary to break the final
task into solvable steps.

\paragraph{Behavior code generation.}
When facing a new curriculum with more units or new unit types, the
behavior coder generates a new script by extending the script that
solved the previous curriculum. The first refinement phase adds new
unit-specific control functions, such as \texttt{control\_vikings} and
\texttt{control\_ghosts}; the second phase refines existing control
functions to improve coordination among units. For example, Marines
initially learn focus fire, and later learn to split against
area-of-effect damage before resuming focus fire.

For further insights into the specific curriculum trajectories, see
Appendix~\ref{app:all_paths_table}. Table~\ref{table:curr} presents the
step-by-step sub-task generation across all 12 scenarios, while
Table~\ref{tab:path_1_only} gives a representative full path showing how
the framework advances through successful nodes and backs off from failed
proposals.

\paragraph{Layered critic refinement.}
Although LLMs learn extensive coding resources during pre-training,
their ability to generate scripts for the \texttt{python-sc2} package
depends on the amount and quality of community-written examples they
have seen. We therefore use a two-layer critic. The first layer is a
sanity-check module that corrects grammar bugs, API misuse, and runtime
exceptions. The second layer refines strategy-level logic in the
behavior-tree script. This two-layer critic improves the rate of
successful behavior-tree generation.

\section{Cross-Domain Experiment Details}
\label{app:crossdomain_details}

This appendix gives the experimental details behind
Table~\ref{tab:crossdomain_summary}. The main text uses these results
only to support the method-level generality claim; the implementation
settings, failure modes, and auxiliary figures are kept here to avoid
interrupting the SC2-led narrative.

\subsection{Flatland}
\label{sec:cd_flatland}

Flatland~\citep{mohanty2020flatland} is a 2D-grid multi-train
pathfinding benchmark. Our final task is a 35$\times$35 grid with
3 cities and 4 trains under \texttt{single\_speed} and
\texttt{no\_malfunctions}; a run is counted as successful when mean
completion is at least $0.90$. Direct LLM generation fails because the
API requires direction-conditioned transitions through
\texttt{env.rail.get\_transitions(r,c,d)}, and valid routing must search
over (cell, direction) rather than cells alone. Multi-train collision
avoidance further requires priority-aware reservations.

Across three random seeds, EvoCurr reaches $0.710 \pm 0.195$ mean
completion, while Direct and Reflexion-5 remain at 0.000 on all seeds.
A more adversarial Reflexion-20 ablation on seed 42 also produces
0/20 generations with non-zero completion despite spending roughly
$3\times$ EvoCurr's training budget.

\subsection{POGEMA}
\label{sec:cd_pogema}

POGEMA~\citep{skrynnik2025pogema} is a partially observable multi-agent
pathfinding benchmark. Our final task is a 20$\times$20 grid with
8 agents, obstacle density 0.25, and observation radius 3. The metric is
ISR (Individual Success Rate), with a success threshold of $0.90$.
Unlike Flatland, POGEMA requires time-indexed reservations and global
belief accumulation from local $7\times7$ observations.

EvoCurr reaches $0.693 \pm 0.108$ mean ISR across three seeds, while
Direct and Reflexion-6 remain at 0.000. In the Reflexion-20 ablation,
17/20 generations crash and the remaining 3/20 execute but deadlock at
ISR$=0$. The most diagnostic failure is repeated
\texttt{NameError: heapq}, which appears on six later generations even
after feedback, showing that generic reflection does not reliably repair
basic executable-structure errors.

\begin{table}[!ht]
  \centering
  \scriptsize
  \caption{Multi-seed verification on the two main supplementary domains.
    Direct one-shot and short-horizon Reflexion are both 0.000 across
    all 3 seeds; \textsc{EvoCurr} is the only non-zero method.
    Reflexion-20 numbers are seed-42 only.}
  \label{tab:flatland_pogema_multiseed}
  \begin{tabular}{l c c c c}
    \toprule
    \textbf{Method} & \textbf{seed 42} & \textbf{seed 43} & \textbf{seed 44} & \textbf{mean $\pm$ s.d.} \\
    \midrule
    \multicolumn{5}{l}{\emph{Flatland -- completion}} \\
    hand\_rule              & 0.17  & 0.17  & 0.00  & 0.113 $\pm$ 0.098 \\
    direct\_oneshot         & 0.000 & 0.000 & 0.000 & 0.000 \\
    Reflexion-5             & 0.000 & 0.000 & 0.000 & 0.000 \\
    Reflexion-20            & 0.000 & --    & --    & 0.000 \\
    \textbf{\textsc{EvoCurr}} & \textbf{0.880} & \textbf{0.750} & \textbf{0.500} & \textbf{0.710 $\pm$ 0.195} \\
    \midrule
    \multicolumn{5}{l}{\emph{POGEMA -- ISR}} \\
    hand\_rule              & 0.04  & --    & 0.08  & 0.060 \\
    direct\_oneshot         & 0.000 & --    & --    & 0.000 \\
    Reflexion-6             & 0.000 & 0.000 & 0.000 & 0.000 \\
    Reflexion-20            & 0.000 & --    & --    & 0.000 (17/20 crash) \\
    \textbf{\textsc{EvoCurr}} & \textbf{0.710} & \textbf{0.580} & \textbf{0.790} & \textbf{0.693 $\pm$ 0.108} \\
    \bottomrule
  \end{tabular}
\end{table}

\subsection{LiveCodeBench Hard}
\label{sec:cd_livecodebench}

We evaluate 20 \emph{Hard}-tier problems from LiveCodeBench
version~v6~\citep{jain2025livecodebench}, restricted to contest dates
after 2024-08 to mitigate contamination relative to the DeepSeek-V4-Pro
training cutoff. Direct one-shot makes one Coder call. Reflexion-3
uses up to three Coder + Summarizer rounds against the full test suite.
EvoCurr uses the same backbone and three-round budget, but decomposes
each problem into a three-stage constraint-relaxation curriculum:
produce runnable code, pass at least one test case, then pass all test
cases. The accepted-floor mechanism prevents later feedback from
destroying an already-working stage.

\begin{table}[!ht]
  \centering
  \tiny
  \caption{LiveCodeBench Hard (post-2024-08, contamination-controlled):
    per-method Pass@1 over 20 problems. All methods use the same
    DeepSeek-V4-Pro backbone.}
  \label{tab:apps_livecodebench}
  \begin{tabular}{l c c c c}
    \toprule
    \textbf{Method} & \textbf{Solved (Pass@1)} & \textbf{Avg pass rate} & \textbf{LLM calls} & \textbf{Cost (USD)} \\
    \midrule
    Direct one-shot           & 8/20 (40\%)  & 40.0\% & 20 & \$0.30 \\
    Reflexion-3               & 8/20 (40\%)  & 40.0\% & 44 & \$0.74 \\
    \textbf{\textsc{EvoCurr}} & \textbf{11/20 (55\%)} & \textbf{58.3\%} & 69 & \$1.03 \\
    \bottomrule
  \end{tabular}
\end{table}

EvoCurr solves 11/20 problems versus 8/20 for Direct and Reflexion-3.
Iteration alone does not close the gap: Reflexion-3 spends
$2.5\times$ the cost of Direct yet obtains the same Pass@1, gaining one
problem while regressing on another (\texttt{abc388\_e}: $100\%\to0\%$).
EvoCurr's gains concentrate on problems where the runnable-code warm-up
stage creates a scaffold that direct generation never reaches.

\subsection{Boundary Controls}
\label{sec:cd_boundary}

\paragraph{pandapipes.}
On an 18-junction gas-network control benchmark, 6-iteration Reflexion
already reaches $1.00$ feasibility and $0.009$ MW power consumption,
outperforming EvoCurr's $0.33$ feasibility. This is a deliberate
negative control: the control space is low-dimensional, the feedback is
direct, and PI-style code templates are within the LLM's one-shot reach.

\paragraph{JSSP.}
On Taillard 10$\times$5 Job-Shop Scheduling, all three methods tie SPT
(Shortest Processing Time) at score $1.00$. The LLM's first attempt is
already SPT-like, so the curriculum cannot exceed an already-saturated
ceiling. Reflexion can even degrade the working solution, consistent
with the Summarizer-drift failure mode observed in LiveCodeBench.

\section{StarCraft~II Micro-Management Details}
\label{sec:appendix}
\subsection{Introduction to StarCraft II}
\label{appendix:introduction_of_starcraft2}

StarCraft~II is a real-time strategy game developed by Blizzard
Entertainment. It has three asymmetric factions, Terran, Protoss, and
Zerg, each with distinct units, technologies, and tactical interactions.
Players must simultaneously manage economy, production, technology,
army composition, and real-time combat control. This makes the game a
long-standing benchmark for studying decision-making under partial
observability, large action spaces, and severe time pressure.

For AI research, StarCraft~II is valuable because it combines
long-horizon strategic planning with millisecond-level tactical
execution. AlphaStar demonstrated that large-scale learning systems can
reach Grandmaster-level full-game play, while PySC2 and SMAC-style
benchmarks made controlled micro-management evaluation more accessible.
Our work focuses on a narrower but difficult slice of this space:
fixed-army late-game engagements where the economy and production
systems are removed, leaving the agent to solve the tactical question of
how to fight with a large heterogeneous army.

\subsection{StarCraft II API and Python Interfaces}

The technical foundation enabling AI research in StarCraft II rests on Blizzard Entertainment's official StarCraft II Machine Learning API, which provides programmatic access to the game's complete state information and action execution capabilities. This API exposes the game engine through a protocol buffer-based interface that delivers real-time observations including unit positions, resource states, map geometry, and tactical information while accepting high-level commands for unit control, building construction, and technology research. The official \textbf{s2client-proto} defines the core communication protocol between external programs and the StarCraft II executable, establishing standardized data structures for observations, actions, and game configuration. This low-level interface handles the complex details of game state serialization, network communication, and command validation, but requires substantial boilerplate code and deep understanding of the underlying protocol specifications to implement effective AI agents.

Building upon this foundation, the research community has developed higher-level abstractions that significantly simplify AI development while preserving the full functionality of the underlying API. \textbf{PySC2}, developed by DeepMind, transforms the raw API into a structured reinforcement learning environment that follows standard RL conventions with observation spaces, action spaces, and reward functions. This environment emphasizes feature-layer representations and provides built-in mini-games for curriculum learning, making it particularly suitable for deep reinforcement learning approaches. 

Complementing PySC2, the \textbf{python-sc2} library offers a more direct and intuitive interface focused on scripted bot development, where complex strategic behaviors can be implemented using straightforward Python code with minimal boilerplate. The python-sc2 library abstracts away protocol buffer complexities while exposing high-level game objects such as units, abilities, and map structures through clean Python APIs, enabling researchers to focus on strategic logic rather than low-level implementation details. Our EvoCurr framework leverages python-sc2's accessibility and expressiveness to generate behavior tree scripts that can be easily interpreted, debugged, and modified, making it an ideal choice for our curriculum-based approach to complex tactical reasoning.

\subsection{Final Task Setting}
\label{app:final_task_setting}
{

\begin{table}[hbt!]
  \centering
  \normalsize 
  \setlength{\tabcolsep}{2pt} 
  
  \caption{Final Terran vs Protoss Task Specification}
  \label{tab:final_task_tvp}
  
  \begin{tabular}{lcl}
    \toprule
    \textbf{Unit Type} & \textbf{Quantity} & \textbf{Technology} \\
    \midrule
    
    \multicolumn{3}{c}{\textit{\textbf{AGENTS (Terran)}}} \\
    \midrule
    Marine & 20 & Stimpack \\
    Marauder & 12 & Stimpack \\
    Medivac & 4 & Heal \\
    Ghost & 8 & PersonalCloaking \\
    SiegeTank & 6 & SiegeTech \\
    VikingFighter & 8 & AssaultMode \\
    Cyclone & 7 & LockOn \\
    WidowMine & 7 & Burrow \\
    Raven & 3 & HunterSeeker \\
    Liberator & 2 & DefenderMode \\
    
    \midrule
    \multicolumn{3}{c}{\textit{\textbf{ENEMIES (Protoss)}}} \\
    \midrule
    Zealot & 15 & Charge \\
    Stalker & 14 & BlinkTech \\
    Sentry & 10 & ForceField \\
    HighTemplar & 8 & PsiStormTech \\
    Colossus & 4 & ExtendedThermalLance \\
    Tempest & 5 & GroundAttack \\
    Disruptor & 4 & PurificationNova \\
    Carrier & 4 & InterceptorLaunch \\
    \bottomrule
  \end{tabular}
\end{table}

\begin{table}[hbt!]
  \centering
  \normalsize
  \setlength{\tabcolsep}{2pt}
  
  \caption{Final Terran vs Zerg Task Specification}
  \label{tab:final_task_tvz}
  
  \begin{tabular}{lcl}
    \toprule
    \textbf{Unit Type} & \textbf{Quantity} & \textbf{Technology} \\
    \midrule
    
    \multicolumn{3}{c}{\textit{\textbf{AGENTS (Terran)}}} \\
    \midrule
    Marine & 20 & Stimpack \\
    Marauder & 12 & Stimpack \\
    Medivac & 4 & Heal \\
    Ghost & 8 & PersonalCloaking \\
    SiegeTank & 6 & SiegeTech \\
    VikingFighter & 8 & AssaultMode \\
    Cyclone & 7 & LockOn \\
    WidowMine & 7 & Burrow \\
    Raven & 3 & HunterSeeker \\
    Liberator & 2 & DefenderMode \\
    
    \midrule
    \multicolumn{3}{c}{\textit{\textbf{ENEMIES (Zerg)}}} \\
    \midrule
    Zergling & 60 & ZerglingMovementSpeed \\
    Baneling & 24 & CentrificalHooks \\
    Roach & 15 & GlialReconstitution \\
    Hydralisk & 10 & HydraliskSpeed \\
    Lurker & 6 & Burrow \\
    Corruptor & 10 & FlyerWeaponsLevel1 \\
    Infestor & 3 & EnergyUpgrade \\
    Viper & 4 & FlyerArmorsLevel1 \\
    Overseer & 3 & FlyerArmorsLevel1 \\
    Queen & 4 & MissileWeaponsLevel1 \\
    Broodlord & 4 & FlyerWeaponsLevel1 \\
    \bottomrule
  \end{tabular}
\end{table}

To rigorously evaluate the framework's capacity for handling high-dimensional state spaces and heterogeneous multi-agent coordination, we construct two terminal scenarios ($T_f$) representing late-game full-technology engagements. As delineated in Tables \ref{tab:final_task_tvp} and \ref{tab:final_task_tvz}, these tasks simulate asymmetric warfare against Protoss and Zerg factions, respectively. The agent is tasked with controlling a complex Terran composition that integrates biological infantry (e.g., Marines, Ghosts) with mechanical support (e.g., Siege Tanks, Liberators). This configuration necessitates the execution of precise micro-management actions, including the timely activation of abilities such as \textit{Stimpack}, \textit{Personal Cloaking}, and \textit{Siege Mode}. Conversely, the adversary configurations are initialized with high-threat counter-units possessing lethal Area-of-Effect (AoE) capabilities, specifically High Templars and Colossi in the Protoss matchup, and Banelings and Broodlords in the Zerg matchup. Consequently, successful task completion demands that the agent exhibit emergent tactical behaviors, such as formation dispersion to mitigate AoE damage and priority targeting to neutralize key enemy assets.
}

\subsection{StarCraft~II Score Tables}
\label{app:sc2_score_tables}

\begin{table*}[t]
    \centering
    \normalsize
    \setlength{\tabcolsep}{2pt} 
    \renewcommand{\arraystretch}{1.1}

    \caption{\textbf{Winning Score (Part 1: Ours).} Comparison of EvoCurr variants.}
    \label{table:winning_part1}

    \begin{tabularx}{\linewidth}{l | X X X}
        \toprule
        \textbf{Task} & \textbf{Evo-DeepSeek} & \textbf{Evo-Claude} & \textbf{Evo-GPT} \\
        \midrule
        bush (TVP)     & \textbf{0.6998} & 0.6359 & 0.6963 \\
        corridor (TVP) & \textbf{0.7146} & 0.6095 & 0.6200 \\
        corner (TVP)   & 0.5515 & 0.6367 & 0.6586 \\
        flat (TVP)     & \textbf{0.6996} & 0.6070 & 0.6295 \\
        main (TVP)     & 0.7277 & 0.6565 & 0.6669 \\
        ramp (TVP)     & 0.7226 & 0.6345 & 0.6811 \\
        \midrule
        bush (TVZ)     & \textbf{0.7322} & 0.6701 & 0.6467 \\
        corridor (TVZ) & \textbf{0.6980} & 0.6497 & 0.5751 \\
        corner (TVZ)   & \textbf{0.8405} & 0.7420 & 0.6206 \\
        flat (TVZ)     & \textbf{0.6263} & 0.6125 & 0.5314 \\
        main (TVZ)     & 0.7017 & \textbf{0.7603} & 0.7531 \\
        ramp (TVZ)     & \textbf{0.7466} & 0.6519 & 0.7064 \\
        \bottomrule
    \end{tabularx}

    \vspace{10pt} 

    \caption*{\textbf{ Part 2: Baselines Comparison.}} 
    
    \normalsize
    \setlength{\tabcolsep}{1.5pt} 
    \begin{tabularx}{\linewidth}{l | X X X X X X X}
        \toprule
        \textbf{Task} & \textbf{DeepSeek} & \textbf{GPT} & \textbf{Claude} & \textbf{Gemini} & \textbf{Claude code} & \textbf{Codex} & \textbf{Cursor} \\
        \midrule
        bush (P)   & 0.33 & 0.43 & 0.35 & 0.49 & 0.19 & 0.27 & 0 \\
        corr (P)   & 0.56 & 0.51 & 0.63 & 0.26 & 0.19 & 0.21 & 0 \\
        corn (P)   & \textbf{0.66} & 0.52 & 0.62 & 0.55 & 0.28 & 0.27 & 0 \\
        flat (P)   & 0.41 & 0.56 & 0.39 & 0.30 & 0.21 & 0.22 & 0 \\
        main (P)   & 0.69 & 0.63 & 0.67 & \textbf{0.77} & 0.34 & 0.40 & 0 \\
        ramp (P)   & 0.45 & 0.48 & \textbf{0.74} & 0.66 & 0.21 & 0.23 & 0 \\
        \midrule
        bush (Z)   & 0.61 & 0.34 & 0.64 & 0.36 & 0.46 & 0.31 & 0 \\
        corr (Z)   & 0.58 & 0.56 & 0.65 & 0.59 & 0.34 & 0.31 & 0 \\
        corn (Z)   & 0.41 & 0.69 & 0.77 & 0.80 & 0.74 & 0.62 & 0 \\
        flat (Z)   & 0.36 & 0.29 & 0.49 & 0.37 & 0.31 & 0.27 & 0 \\
        main (Z)   & 0.68 & 0.43 & 0.43 & 0.66 & 0.45 & 0.45 & 0 \\
        ramp (Z)   & 0.45 & 0.45 & 0.68 & 0.29 & 0.45 & 0.29 & 0 \\
        \bottomrule
    \end{tabularx}
    \label{table:winning_Score_direct}
\end{table*}

\begin{table*}[t]
    \centering
    \normalsize
    \setlength{\tabcolsep}{2pt} 
    \renewcommand{\arraystretch}{1.1}

    \caption{\textbf{Reflect Score (Part 1: Ours).} Comparison of EvoCurr variants with self-reflection.}
    \label{table:reflect_part1}

    \begin{tabularx}{\linewidth}{l | X X X}
        \toprule
        \textbf{Task} & \textbf{Evo-DeepSeek} & \textbf{Evo-Claude} & \textbf{Evo-GPT} \\
        \midrule
        pvt+bush     & \textbf{0.6998} & 0.6359 & 0.6963 \\
        pvt+corridor & \textbf{0.7146} & 0.6095 & 0.6200 \\
        pvt+corner   & 0.5515 & 0.6367 & 0.6586 \\
        pvt+flat     & \textbf{0.6996} & 0.6070 & 0.6295 \\
        pvt+main     & 0.7277 & 0.6565 & 0.6669 \\
        pvt+ramp     & \textbf{0.7226} & 0.6345 & 0.6811 \\
        \midrule
        zvt+bush     & \textbf{0.7322} & 0.6701 & 0.6467 \\
        zvt+corridor & \textbf{0.6980} & 0.6497 & 0.5751 \\
        zvt+corner   & \textbf{0.8405} & 0.7420 & 0.6206 \\
        zvt+flat     & \textbf{0.6263} & 0.6125 & 0.5314 \\
        zvt+main     & 0.7017 & \textbf{0.7603} & 0.7531 \\
        zvt+ramp     & \textbf{0.7466} & 0.6519 & 0.7064 \\
        \bottomrule
    \end{tabularx}

    \vspace{10pt} 

    \caption*{\textbf{Part 2: Baselines Comparison.}} 
    
    \normalsize 
    \setlength{\tabcolsep}{2pt} 
    \begin{tabularx}{\linewidth}{l | X X X X}
        \toprule
        \textbf{Task} & \textbf{DeepSeek} & \textbf{GPT-5} & \textbf{Claude-4} & \textbf{Gemini} \\
        \midrule
        pvt+bush     & 0.4605 & 0.1667 & 0.5208 & 0 \\
        pvt+corridor & 0.3328 & 0.2649 & 0.4651 & 0 \\
        pvt+corner   & \textbf{0.7332} & 0.0899 & 0.6823 & 0 \\
        pvt+flat     & 0.5467 & 0.4502 & 0.6110 & 0 \\
        pvt+main     & 0.6518 & 0.4196 & \textbf{0.7381} & 0 \\
        pvt+ramp     & 0.4679 & 0.4142 & 0.5935 & 0.6649 \\
        \midrule
        zvt+bush     & 0.3189 & 0.2134 & 0.4271 & 0 \\
        zvt+corridor & 0.5696 & 0.4746 & 0.2913 & 0.4614 \\
        zvt+corner   & 0.7043 & 0.7620 & 0.5773 & 0 \\
        zvt+flat     & 0.4520 & 0.2765 & 0.3424 & 0 \\
        zvt+main     & 0.3068 & 0.2982 & 0.5518 & 0.4567 \\
        zvt+ramp     & 0.2691 & 0.5488 & 0.3305 & 0 \\
        \bottomrule
    \end{tabularx}
    \label{table:reflect_score_comparison}
\end{table*}

Win rate captures final success but not the quality of partial or already-winning policies. We therefore also report a damage-cost-aware combat score $\mathcal{S}_{combat}$ for more nuanced evaluation. This metric compares EvoCurr with direct target-task implementations generated by closed-source LLMs (DeepSeek-V3.1, GPT-5, Claude-4, and Gemini-2.5), given by:
\begin{equation}
\mathcal{S}_{combat} = 0.5 \cdot \frac{R_{\text{agent\_final}}}{R_{\text{agent\_init}}} + 0.5 \cdot \left(1 - \frac{R_{\text{enemy\_final}}}{R_{\text{enemy\_init}}}\right),
\end{equation}
where the total combat power for one side is
\begin{equation}
R_{\text{side}} = \sum_i (m_i + \alpha v_i + \beta b_i)
\,\frac{h_i + s_i}{h^{\max}_i + s^{\max}_i}.
\end{equation}
This metric aggregates resource cost and remaining health/shields; scores range from $0$ to $1$. A combat score $\mathcal{S}_{combat} > 0.5$ indicates successful annihilation of the majority of enemy forces while preserving our own, providing a finer-grained assessment than binary win/loss especially for failed code executions where the first term becomes 0.

Following the protocol in Section~\ref{sec:exp_sc2}, we keep selected-policy
comparison and cheap-sampling variability separate. EvoCurr cells below are
the final policies selected by complete curriculum searches. Direct selected
cells are the best executable policies found by repeated one-shot target-task
sampling under the same per-policy rollout evaluation. We then report the
direct-attempt distributions separately, since those 10 attempts estimate
one-shot generation variability rather than full curriculum-search
variability.

\begin{table*}[!ht]
  \centering
  \small
  \setlength{\tabcolsep}{5pt}
  \renewcommand{\arraystretch}{1.15}
  \caption{Selected SC2 combat scores for Terran-vs-Protoss tasks.
  EvoCurr columns report the curriculum-selected final policy for each
  backbone. Direct columns report the best executable policy found by
  each direct-generation backbone within 10 target-task attempts.}
  \label{tab:sc2_selected_scores_tvp}
  \begin{tabular}{l ccc cccc}
    \toprule
    & \multicolumn{3}{c}{\textbf{\textsc{EvoCurr} selected}} &
      \multicolumn{4}{c}{\textbf{Direct selected best}} \\
    \cmidrule(lr){2-4} \cmidrule(lr){5-8}
    \textbf{Task} & DS & Claude & GPT &
    DS & GPT & Claude & Gemini \\
    \midrule
    Bush & 0.700 & 0.636 & 0.696 & 0.373 & 0.406 & 0.590 & 0.514 \\
    Corridor & 0.715 & 0.610 & 0.620 & 0.627 & 0.617 & 0.644 & 0.340 \\
    Corner & 0.551 & 0.637 & 0.659 & 0.640 & 0.638 & 0.708 & 0.559 \\
    Flat & 0.700 & 0.607 & 0.629 & 0.433 & 0.652 & 0.331 & 0.319 \\
    Main & 0.728 & 0.657 & 0.667 & 0.666 & 0.706 & 0.766 & 0.841 \\
    Ramp & 0.723 & 0.635 & 0.681 & 0.658 & 0.550 & 0.766 & 0.631 \\
    \bottomrule
  \end{tabular}
\end{table*}

\begin{table*}[!ht]
  \centering
  \small
  \setlength{\tabcolsep}{6pt}
  \renewcommand{\arraystretch}{1.15}
  \caption{Direct-generation combat-score distributions for
  Terran-vs-Protoss tasks. Cells report mean $\pm$ std over 10 independent
  target-task attempts; brackets give executable attempts out of 10.}
  \label{tab:sc2_direct_distribution_tvp}
  \begin{tabular}{l cccc}
    \toprule
    \textbf{Task} & \textbf{DeepSeek-V3.1} & \textbf{GPT-5} &
    \textbf{Claude-4} & \textbf{Gemini-2.5} \\
    \midrule
    Bush & 0.098$\pm$0.159 [3] & 0.261$\pm$0.147 [8] & 0.097$\pm$0.210 [2] & 0.212$\pm$0.210 [6] \\
    Corridor & 0.320$\pm$0.242 [7] & 0.259$\pm$0.205 [7] & 0.174$\pm$0.241 [4] & 0.081$\pm$0.118 [4] \\
    Corner & 0.200$\pm$0.278 [4] & 0.413$\pm$0.177 [9] & 0.071$\pm$0.224 [1] & 0.197$\pm$0.222 [6] \\
    Flat & 0.108$\pm$0.177 [3] & 0.238$\pm$0.265 [5] & 0.033$\pm$0.105 [1] & 0.032$\pm$0.101 [1] \\
    Main & 0.247$\pm$0.279 [5] & 0.476$\pm$0.276 [8] & 0.150$\pm$0.316 [2] & 0.084$\pm$0.266 [1] \\
    Ramp & 0.302$\pm$0.279 [6] & 0.197$\pm$0.218 [5] & 0.310$\pm$0.335 [5] & 0.063$\pm$0.199 [1] \\
    \bottomrule
  \end{tabular}
\end{table*}

\begin{table*}[!ht]
  \centering
  \small
  \setlength{\tabcolsep}{5pt}
  \renewcommand{\arraystretch}{1.15}
  \caption{Selected SC2 combat scores for Terran-vs-Zerg tasks.
  The reporting convention matches Table~\ref{tab:sc2_selected_scores_tvp}.}
  \label{tab:sc2_selected_scores_tvz}
  \begin{tabular}{l ccc cccc}
    \toprule
    & \multicolumn{3}{c}{\textbf{\textsc{EvoCurr} selected}} &
      \multicolumn{4}{c}{\textbf{Direct selected best}} \\
    \cmidrule(lr){2-4} \cmidrule(lr){5-8}
    \textbf{Task} & DS & Claude & GPT &
    DS & GPT & Claude & Gemini \\
    \midrule
    Bush & 0.732 & 0.670 & 0.647 & 0.646 & 0.430 & 0.661 & 0.357 \\
    Corridor & 0.698 & 0.650 & 0.575 & 0.668 & 0.541 & 0.676 & 0.675 \\
    Corner & 0.840 & 0.742 & 0.621 & 0.495 & 0.745 & 0.764 & 0.789 \\
    Flat & 0.626 & 0.612 & 0.531 & 0.362 & 0.318 & 0.548 & 0.353 \\
    Main & 0.702 & 0.760 & 0.753 & 0.786 & 0.535 & 0.661 & 0.657 \\
    Ramp & 0.747 & 0.652 & 0.706 & 0.428 & 0.411 & 0.681 & 0.329 \\
    \bottomrule
  \end{tabular}
\end{table*}

\begin{table*}[!ht]
  \centering
  \small
  \setlength{\tabcolsep}{6pt}
  \renewcommand{\arraystretch}{1.15}
  \caption{Direct-generation combat-score distributions for
  Terran-vs-Zerg tasks. The reporting convention matches
  Table~\ref{tab:sc2_direct_distribution_tvp}.}
  \label{tab:sc2_direct_distribution_tvz}
  \begin{tabular}{l cccc}
    \toprule
    \textbf{Task} & \textbf{DeepSeek-V3.1} & \textbf{GPT-5} &
    \textbf{Claude-4} & \textbf{Gemini-2.5} \\
    \midrule
    Bush & 0.255$\pm$0.283 [5] & 0.226$\pm$0.196 [6] & 0.100$\pm$0.224 [2] & 0.071$\pm$0.149 [2] \\
    Corridor & 0.325$\pm$0.168 [9] & 0.171$\pm$0.225 [4] & 0.068$\pm$0.214 [1] & 0.130$\pm$0.273 [2] \\
    Corner & 0.121$\pm$0.201 [3] & 0.356$\pm$0.324 [6] & 0.288$\pm$0.373 [4] & 0.265$\pm$0.347 [4] \\
    Flat & 0.123$\pm$0.161 [4] & 0.149$\pm$0.158 [5] & 0.144$\pm$0.200 [4] & 0.035$\pm$0.112 [1] \\
    Main & 0.534$\pm$0.240 [9] & 0.297$\pm$0.213 [7] & 0.066$\pm$0.209 [1] & 0.234$\pm$0.309 [4] \\
    Ramp & 0.276$\pm$0.192 [7] & 0.227$\pm$0.166 [7] & 0.124$\pm$0.228 [3] & 0.065$\pm$0.137 [2] \\
    \bottomrule
  \end{tabular}
\end{table*}

To demonstrate the effectiveness of EvoCurr, we compared it against direct behavior tree generation using LLMs and an approach that enables models to refine behavior trees through self-reflection. The results are presented in Tables \ref{table:winning_Score_direct} and \ref{table:reflect_score_comparison}.

\clearpage

\subsection{Generated Behavior Tree Code Examples}

This section presents complete examples of behavior tree implementations generated by the EvoCurr framework at different curriculum stages, demonstrating the evolution of tactical complexity.

\subsubsection{Early Stage: Basic Marine Micro-management}
{
\begin{figure}[!ht]
\centering
\begin{lstlisting}[
    style=mystyle,
    language=Python,
    basicstyle=\ttfamily\scriptsize,
    numbers=left,
    breaklines=true,
    frame=lines,
    framesep=2mm,
    xleftmargin=2em,
    numbersep=4pt
]
from sc2 import maps
from sc2.bot_ai import BotAI
from sc2.data import Race, Difficulty
from sc2.ids.ability_id import AbilityId
from sc2.ids.unit_typeid import UnitTypeId
from sc2.main import run_game
from sc2.player import Bot, Computer

class BattleBot(BotAI):
    def __init__(self):
        super().__init__()
        self.stim_used = set()

    async def on_step(self, iteration: int):
        if iteration == 0:
            print("Marine Micro Bot - 5v2 Marines vs Zealots!")
        if self.units.exists:
            await self.marine_micro()

    async def marine_micro(self):
        marines = self.units(UnitTypeId.MARINE)
        zealots = self.enemy_units(UnitTypeId.ZEALOT)

        if not marines.exists or not zealots.exists:
            return

        close_zealots = zealots.filter(lambda z:
            marines.closest_to(z.position).distance_to(z) < 6)

        if close_zealots.exists:
            for marine in marines:
                if (marine.tag not in self.stim_used and
                    AbilityId.EFFECT_STIM_MARINE in 
                    await self.get_available_abilities(marine)):
                    marine(AbilityId.EFFECT_STIM_MARINE)
                    self.stim_used.add(marine.tag)

        target = min(zealots, key=lambda z: z.health + z.shield)

        for marine in marines:
            dist = marine.distance_to(target)
            if dist < 1:
                retreat_pos = marine.position.towards(target.position, -3)
                marine.move(retreat_pos)
            elif dist <= 5:
                marine.attack(target)
            else:
                marine.move(target.position)
\end{lstlisting}
\end{figure}
}

\subsubsection{Intermediate Stage: Multi-unit Coordination}
{
\normalsize
\label{code:intermediate}
\Needspace{18\baselineskip}
\begin{figure}[!ht]
\centering
\begin{lstlisting}[
    style=mystyle,
    language=Python,
    basicstyle=\ttfamily\scriptsize,
    numbers=left,
    breaklines=true,
    frame=lines,
    framesep=2mm,
    xleftmargin=2em,
    numbersep=4pt
]
class BattleBot(BotAI):
    async def on_step(self, iteration: int):
        if iteration == 0:
            print("Terran Battle Bot Activated!")
        
        if self.units.exists:
            await self.control_ghosts()
            await self.control_marines()
            await self.control_marauders()
            await self.control_medivacs()

    async def control_ghosts(self):
        ghosts = self.units(UnitTypeId.GHOST)
        if not ghosts.exists:
            return
            
        high_templars = self.enemy_units(UnitTypeId.HIGHTEMPLAR)
        stalkers = self.enemy_units(UnitTypeId.STALKER)
        
        for ghost in ghosts:
            if high_templars.exists:
                templar = high_templars.closest_to(ghost.position)
                if ghost.distance_to(templar) < 12:
                    if AbilityId.SNIPE_SNIPE in 
                       await self.get_available_abilities(ghost):
                        ghost(AbilityId.SNIPE_SNIPE, templar)
                    elif AbilityId.EMP_EMP in 
                         await self.get_available_abilities(ghost):
                        ghost(AbilityId.EMP_EMP, templar.position)
                ghost.move(templar.position)
                if AbilityId.BEHAVIOR_CLOAKON_GHOST in 
                   await self.get_available_abilities(ghost):
                    ghost(AbilityId.BEHAVIOR_CLOAKON_GHOST)
            elif stalkers.exists:
                target = stalkers.closest_to(ghost.position)
                ghost.attack(target)

    async def control_medivacs(self):
        medivacs = self.units(UnitTypeId.MEDIVAC)
        if not medivacs.exists:
            return
            
        bio_units = self.units.filter(lambda unit: 
                    unit.type_id in {UnitTypeId.MARINE, 
                    UnitTypeId.MARAUDER})
        for medivac in medivacs:
            injured = bio_units.filter(lambda unit: 
                      unit.health_percentage < 0.75)
            if injured.exists:
                target = injured.closest_to(medivac.position)
                medivac.move(target.position)
                if medivac.distance_to(target) < 5:
                    medivac(AbilityId.MEDIVACHEAL_HEAL, target)
            elif bio_units.exists:
                medivac.move(bio_units.center)
\end{lstlisting}
\end{figure}
}

\subsubsection{Advanced Stage: Complex Multi-unit Tactical Framework}
{
\label{code:advanced}
\normalsize
\Needspace{18\baselineskip}
\begin{figure}[!ht]
\centering
\begin{lstlisting}[
    style=mystyle,
    language=Python,
    basicstyle=\ttfamily\scriptsize,
    numbers=left,
    breaklines=true,
    frame=lines,
    framesep=2mm,
    xleftmargin=2em,
    numbersep=4pt
]

class BattleBot(BotAI):
    async def on_step(self, iteration: int):
        if iteration == 0:
            self.setup_complete = False
            await self.initial_positioning()
            self.setup_complete = True
        
        bio_units = self.units.of_type({UnitTypeId.MARINE, 
        UnitTypeId.MARAUDER})
        medivacs = self.units(UnitTypeId.MEDIVAC)

        await self.avoid_aoe(bio_units + medivacs)
        await self.control_siege_tanks()
        await self.control_vikings()
        await self.control_liberators()
        await self.control_ghosts()
        await self.control_bio(UnitTypeId.MARINE)
        await self.control_bio(UnitTypeId.MARAUDER)
        await self.control_medivacs()

    async def avoid_aoe(self, units: Units):
        storms = [e for e in self.state.effects 
                 if e.id == EffectId.PSISTORMPERSISTENT]
        disruptor_balls = self.enemy_units(UnitTypeId.DISRUPTORPHASED)
        
        threats = []
        for storm in storms:
            threats.append((storm.position, 2.5)) 
        for ball in disruptor_balls:
            threats.append((ball.position, 2.5))

        for unit in units:
            for pos, radius in threats:
                if unit.distance_to(pos) < radius:
                    away = unit.position.towards(pos, -3)
                    unit.move(away)
                    break

    async def control_siege_tanks(self):
        siege_tanks = self.units(UnitTypeId.SIEGETANKSIEGED)
        sieged_tanks = self.units(UnitTypeId.SIEGETANKSIEGED)

        for tank in siege_tanks:
            if tank.distance_to(Point2((15, 15))) > 2:
                continue
            abilities = await self.safe_get_abilities(tank)
            if AbilityId.SIEGEMODE_SIEGEMODE in abilities:
                tank(AbilityId.SIEGEMODE_SIEGEMODE)
        
        if sieged_tanks.exists:
            enemies = self.enemy_units
            if not enemies.exists:
                return
            priority_targets = enemies.of_type([UnitTypeId.COLOSSUS, 
                              UnitTypeId.STALKER, UnitTypeId.HIGHTEMPLAR, 
                              UnitTypeId.ZEALOT])
            
            for tank in sieged_tanks:
                targets_in_range = priority_targets.in_attack_range_of(tank)
                if targets_in_range:
                    target = min(targets_in_range, 
                            key=lambda t: (t.type_id not in 
                            {UnitTypeId.COLOSSUS, UnitTypeId.HIGHTEMPLAR}, 
                            t.distance_to(tank)))
                    tank.attack(target)
\end{lstlisting}
\end{figure}
}

\section{Complete Curriculum Evolution Paths (StarCraft~II)}
\label{app:all_paths_table}

\subsection{Curriculum Design} 

\begin{figure*}[t]
    \centering
    \includegraphics[trim=0cm 2cm 0cm 1.5cm, clip, width=1\linewidth]{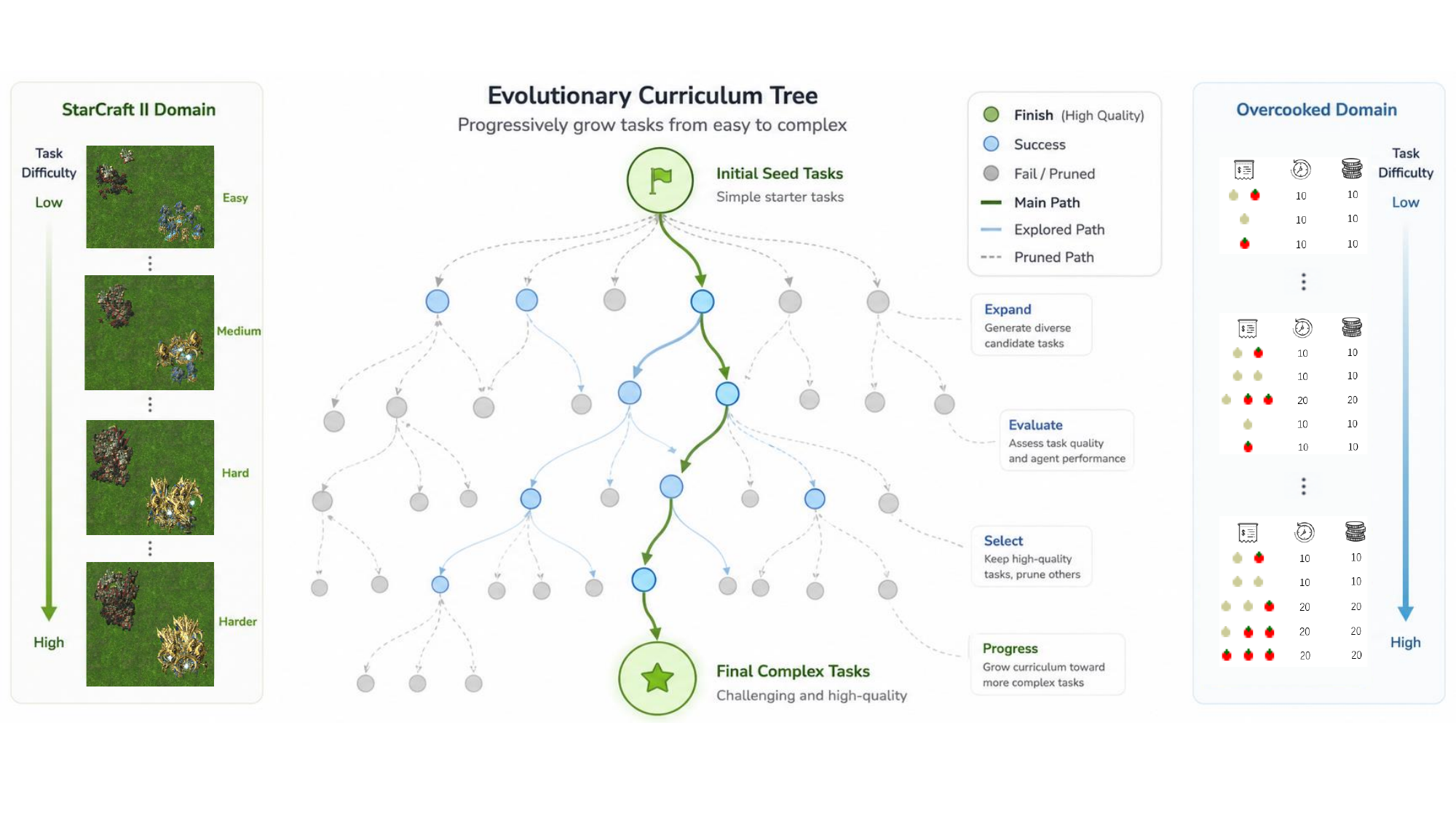}
    \caption{Conceptual illustration of the self-evolving curriculum tree in EvoCurr. The framework progressively grows task difficulty from simple starter seeds to the final high-quality complex tasks. By enforcing an accepted-floor constraint, the model efficiently prunes failed sub-paths and backs off to mastered checkpoints, preventing policy regression. Representative intermediate states from StarCraft II and Overcooked domains are illustrated on both sides.}
    \label{fig:evolution}
\end{figure*}

\begin{table*}[t]
    \centering
    \normalsize 
    \setlength{\tabcolsep}{1pt} 
    \renewcommand{\arraystretch}{1.2}

    \newcolumntype{Y}{>{\centering\arraybackslash}X}

    \caption{Winning rates (\%) of curricula designed for the 12 complex decision-making scenarios.}
    \label{table:curr}

    \begin{tabularx}{\textwidth}{@{} l *{7}{Y} @{}}
        \toprule
        \textbf{Task} & \textbf{Task1} & \textbf{Task2} & \textbf{Task3} & \textbf{Task4} & \textbf{Task5} & \textbf{Task6} & \textbf{Task7} \\
        \midrule
        Flat (TvP)     & 100 & 100 & 70  & 90  & 100 & -   & - \\ 
        Flat (TvZ)     & 100 & 100 & 100 & 100 & 100 & -   & - \\ 
        Bush (TvP)     & 100 & 100 & 40  & 90  & 50  & 100 & 100 \\ 
        Bush (TvZ)     & 100 & 90  & 60  & 100 & 100 & -   & - \\ 
        Corridor (TvP) & 100 & 90  & 100 & 100 & 90  & -   & - \\ 
        Corridor (TvZ) & 100 & 90  & 90  & 100 & 100 & -   & - \\ 
        Corner (TvP)   & 90  & 90  & 90  & 100 & -   & -   & - \\ 
        Corner (TvZ)   & 100 & 100 & 90  & 90  & 100 & -   & - \\ 
        Main (TvP)     & 100 & 100 & 100 & 100 & -   & -   & - \\ 
        Main (TvZ)     & 100 & 100 & 100 & 100 & -   & -   & - \\
        Ramp (TvP)     & 100 & 100 & 90  & 100 & 100 & -   & - \\
        Ramp (TvZ)     & 100 & 100 & 100 & 90  & 100 & -   & - \\
        \bottomrule
    \end{tabularx}
\end{table*}

We illustrate the curriculum-design process by listing the sub-tasks generated for the final task in Table~\ref{table:curr}. Each sub-task is generated based on the outcome of the previous curriculum. For StarCraft~II tasks, we set the acceptance threshold to $\tau = 0.9$: the Designer increases difficulty when the win rate is above 90\% and decreases difficulty otherwise. Among the 12 tasks, the Terran vs Protoss setting on the Bush map requires the longest curriculum trajectory because bush concealment makes enemy visibility and target selection harder; the Designer therefore lowers difficulty twice before the Solver finally masters the target task.

We omit the older evolution schematic here because it is visually noisy in
the two-column appendix. The main paper keeps the clearer trajectory view
in Figure~\ref{fig:evolution}; this appendix focuses on the exact
curriculum tables needed for auditability.

\subsection{Example EvoCurr Evolution Path}

\begin{table*}[t]
  \centering
  \small 
  \setlength{\tabcolsep}{3pt} 
  \renewcommand{\arraystretch}{1.35} 
  
  \caption{\textbf{Curriculum Evolution Trace (Path 1).} Detailed breakdown of unit compositions from the initial task to the final complex scenario.}
  \label{tab:path_1_only}
  
  \newcolumntype{L}{>{\raggedright\arraybackslash}X}
  
  \begin{tabularx}{\textwidth}{c L c}
    \toprule
    \textbf{Stage} & \textbf{Composition (Agent vs Enemy)} & \textbf{Win Rate} \\
    \midrule
    
    1 & \textbf{Agent:} Marine (5) \newline 
        \textbf{Enemy:} Zealot (2, Charge) & 90\% \\ 
    \midrule
    
    2 & \textbf{Agent:} Marine (10), Marauder (5), Medivac (1) \newline 
        \textbf{Enemy:} Zealot (5, Charge), Stalker (5, Blink), HighTemplar (1, PsiStorm) & 90\% \\ 
    \midrule
    
    3 & \textbf{Agent:} Marine (15), Marauder (8), Ghost (2), Medivac (2), SiegeTank (1) \newline 
        \textbf{Enemy:} Zealot (10, Charge), Stalker (8, Blink), HighTemplar (2, PsiStorm), Colossus (1, ExtendedThermalLance) & Failed \\ 
    \midrule
    
    4 & \textbf{Agent:} Marine (12), Marauder (6), Ghost (1), Medivac (1) \newline 
        \textbf{Enemy:} Zealot (8, Charge), Stalker (6, Blink), HighTemplar (1, PsiStorm) & 90\% \\ 
    \midrule
    
    5 & \textbf{Agent:} Marine (18), Marauder (10), Ghost (2), Medivac (2), SiegeTank (1), Viking (2) \newline 
        \textbf{Enemy:} Zealot (12, Charge), Stalker (10, Blink), HighTemplar (2, PsiStorm), Colossus (1, ExtendedThermalLance) & 90\% \\ 
    \midrule
    
    6 & \textbf{Agent:} Marine (20), Marauder (12), Medivac (4), Ghost (8), SiegeTank (6), VikingFighter (8), Cyclone (7), WidowMine (7), Raven (3), Liberator (2) \newline 
        \vspace{2pt} 
        \textbf{Enemy:} Zealot (15, Charge), Stalker (14, BlinkTech), Sentry (10, ForceField), HighTemplar (8, PsiStormTech), Colossus (4, ExtendedThermalLance), Tempest (5, GroundAttack), Disruptor (4, PurificationNova), Carrier (4, InterceptorLaunch) & 100\% \\
        
    \bottomrule
  \end{tabularx}
\end{table*}

To provide a granular view of the curriculum evolution process, Table~\ref{tab:path_1_only} traces a representative curriculum path in the Terran vs. Protoss scenario. The progression demonstrates the adaptive Designer--Solver dynamics: beginning with a simple engagement of 5 Marines against 2 Zealots, the curriculum escalates through increasingly complex multi-unit compositions incorporating specialized roles such as Ghosts for detection and Siege Tanks for area control. When the Solver fails to achieve the acceptance threshold at Stage 3 with an ambitious 5-unit-type composition, the framework retreats to a simplified 4-unit configuration in Stage 4, re-establishing a 90\% win rate before advancing again. This adaptive adjustment enables the system to reach the final task at Stage 6, where the agent commands a diverse 10-unit-type army against an 8-unit-type enemy force.

\subsection{StarCraft~II Mini-Game Maps}

\begin{figure*}[t]
    \centering
    \begin{minipage}[t]{0.47\textwidth}
        \centering
        \includegraphics[width=0.78\textwidth]{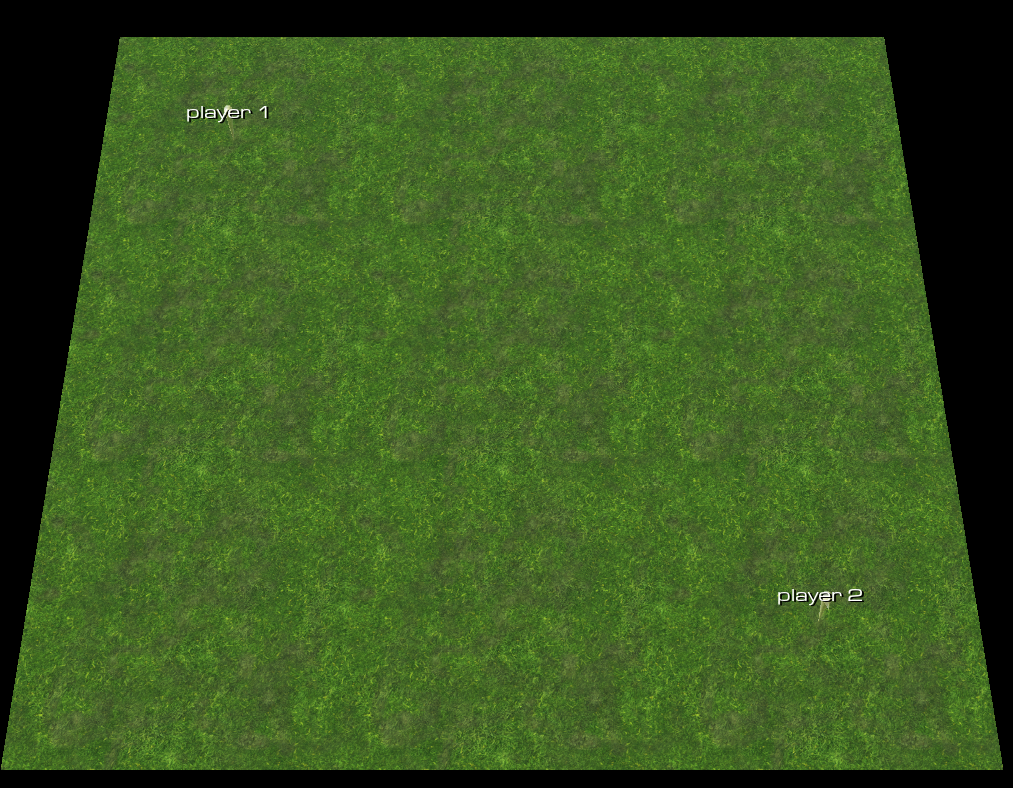}
        \vspace{-0.2cm}
        \caption{Flat map layout}
        \label{fig:flat}
    \end{minipage}
    \hfill
    \begin{minipage}[t]{0.47\textwidth}
        \centering
        \includegraphics[width=0.78\textwidth]{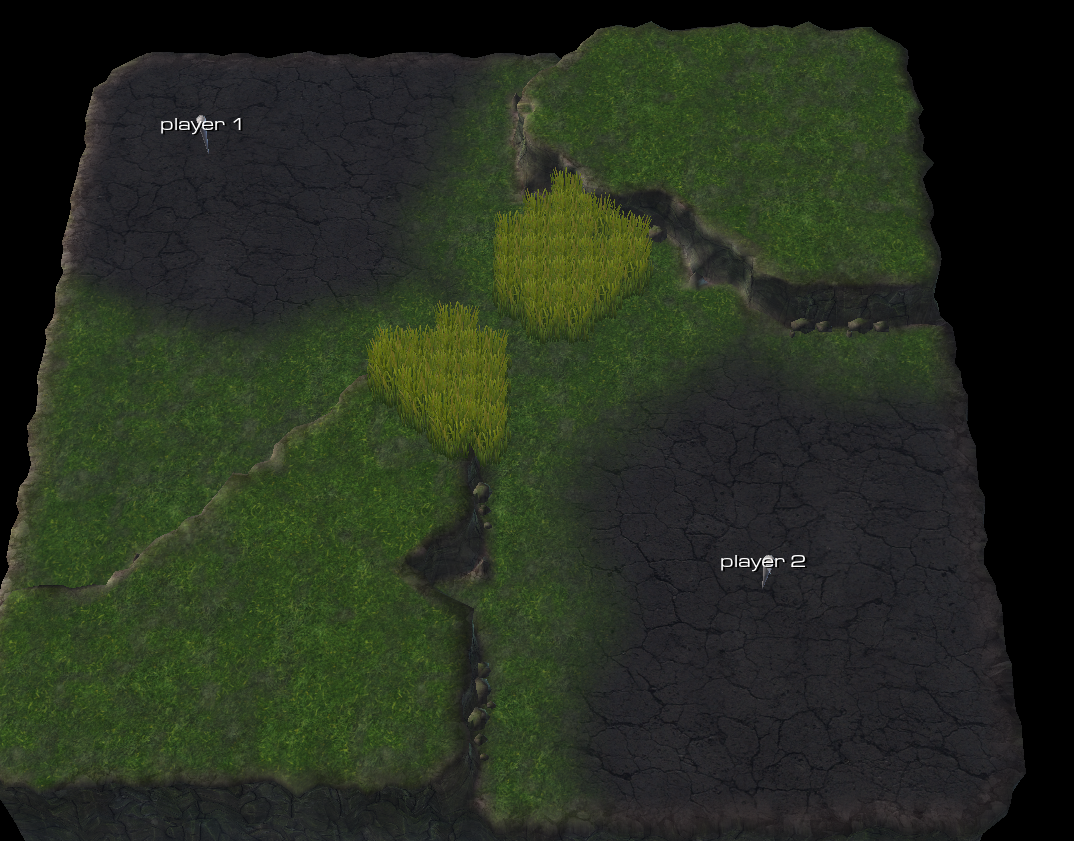}
        \vspace{-0.2cm}
        \caption{Bush map layout}
        \label{fig:bush}
    \end{minipage}
    
    \vspace{0.12cm}
    
    \begin{minipage}[t]{0.47\textwidth}
        \centering
        \includegraphics[width=0.78\textwidth]{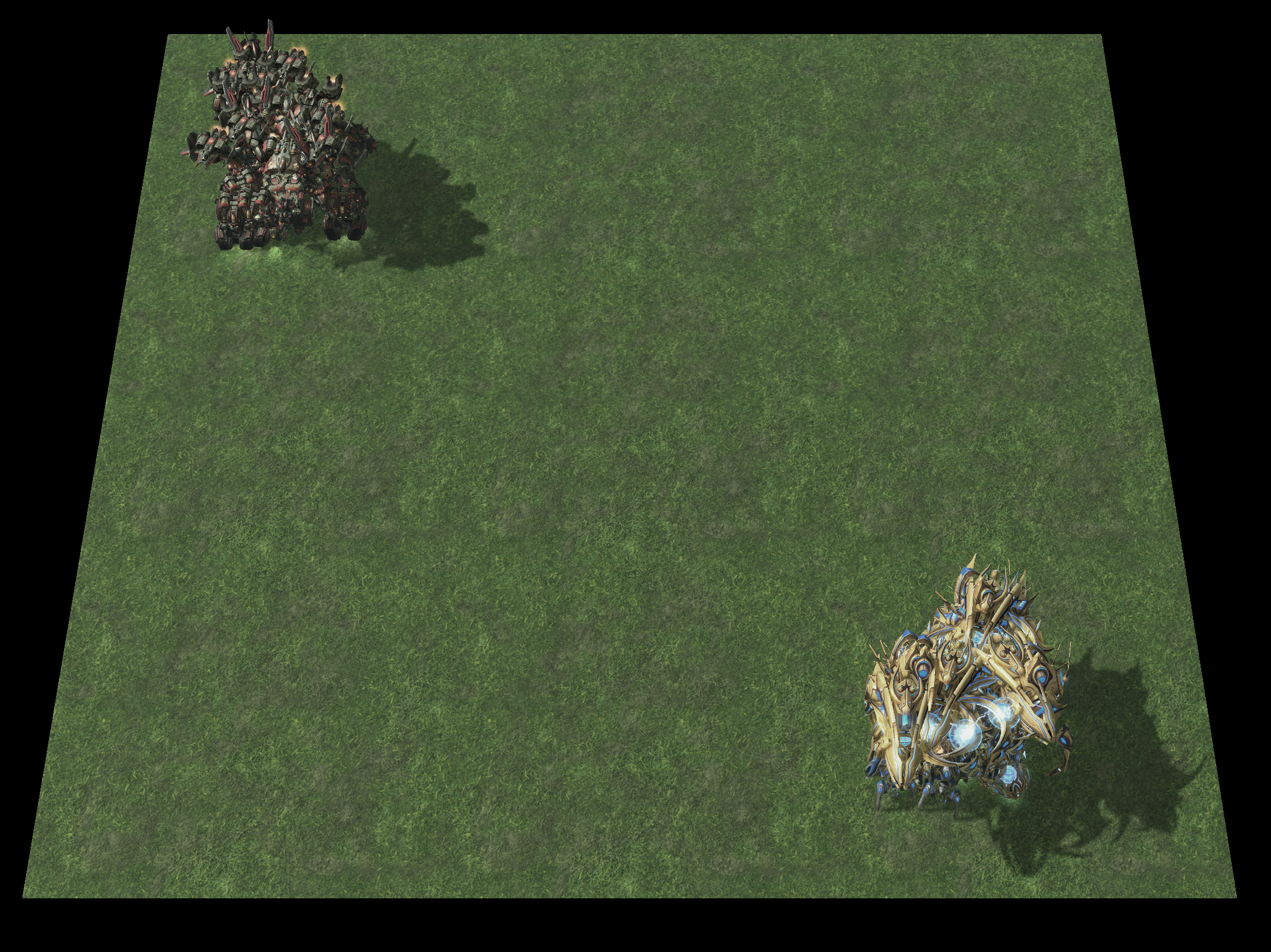}
        \vspace{-0.2cm}
        \caption{Terran vs Protoss on Flat}
        \label{fig:tvp_flat}
    \end{minipage}
    \hfill
    \begin{minipage}[t]{0.47\textwidth}
        \centering
        \includegraphics[width=0.78\textwidth]{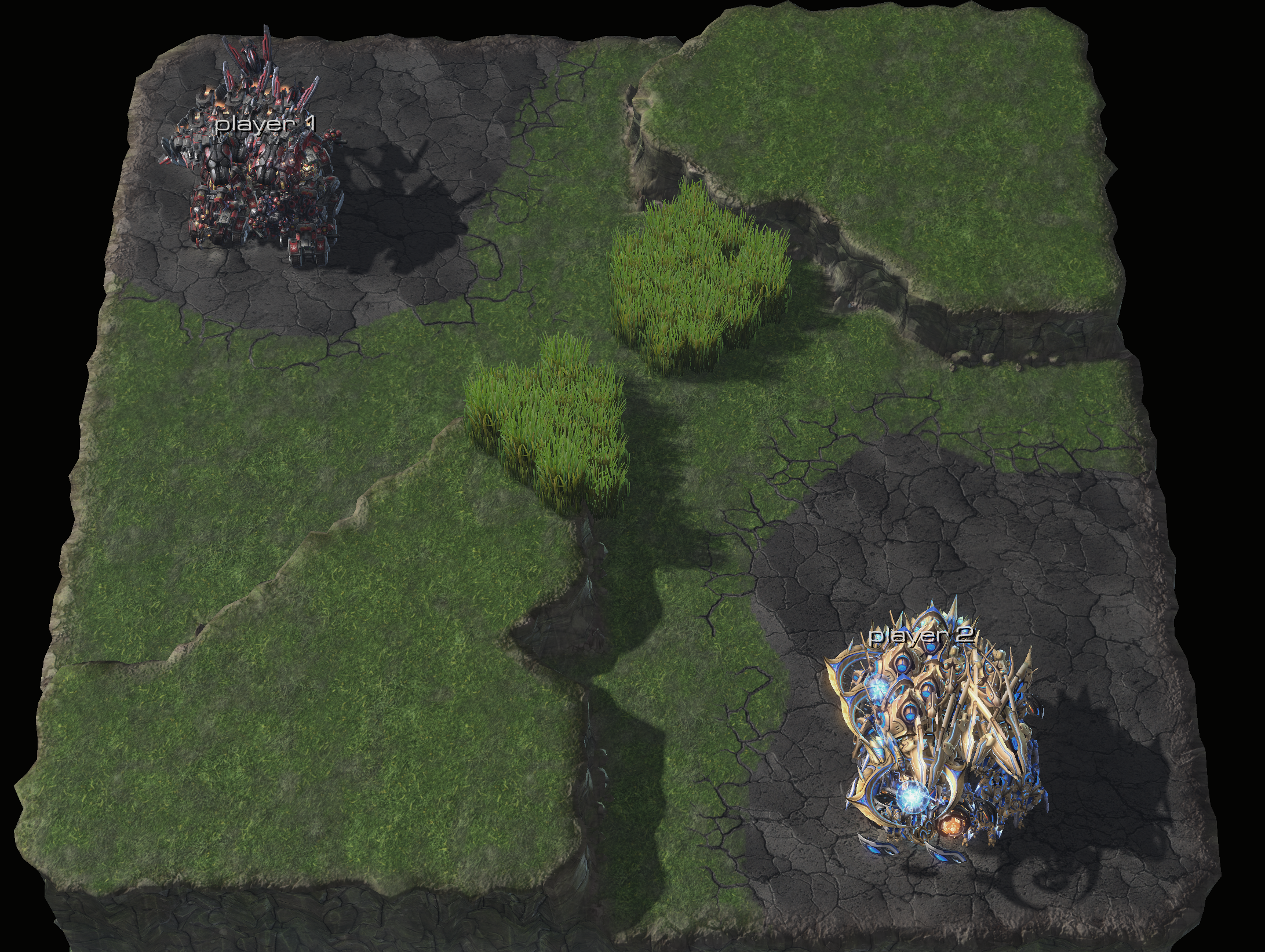}
        \vspace{-0.2cm}
        \caption{Terran vs Protoss on Bush}
        \label{fig:pvt_bush}
    \end{minipage}
    
    \vspace{0.12cm}
    
    \begin{minipage}[t]{0.47\textwidth}
        \centering
        \includegraphics[width=0.78\textwidth]{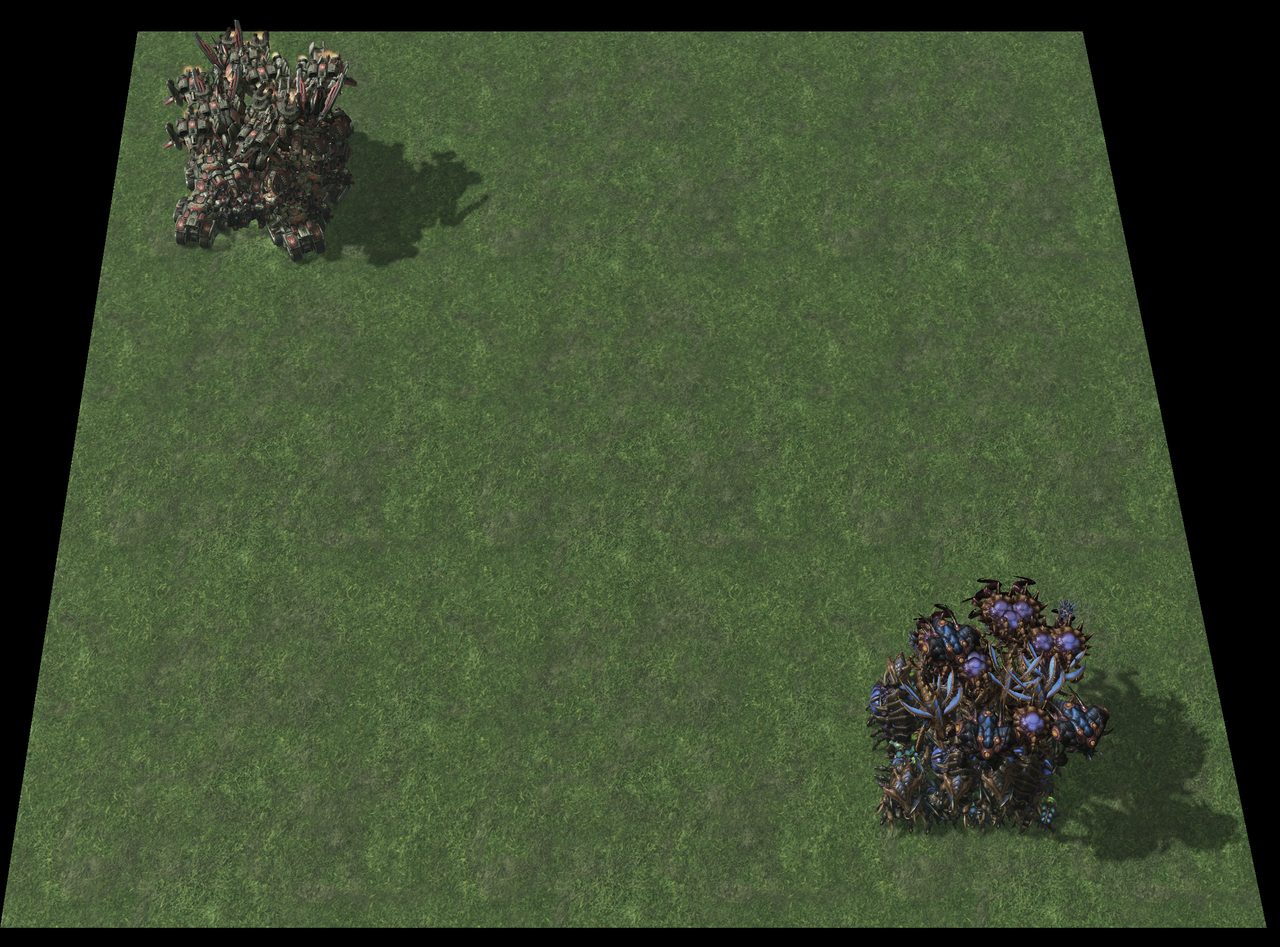}
        \vspace{-0.2cm}
        \caption{Terran vs Zerg on Flat}
        \label{fig:tvz_flat}
    \end{minipage}
    \hfill
    \begin{minipage}[t]{0.47\textwidth}
        \centering
        \includegraphics[width=0.78\textwidth]{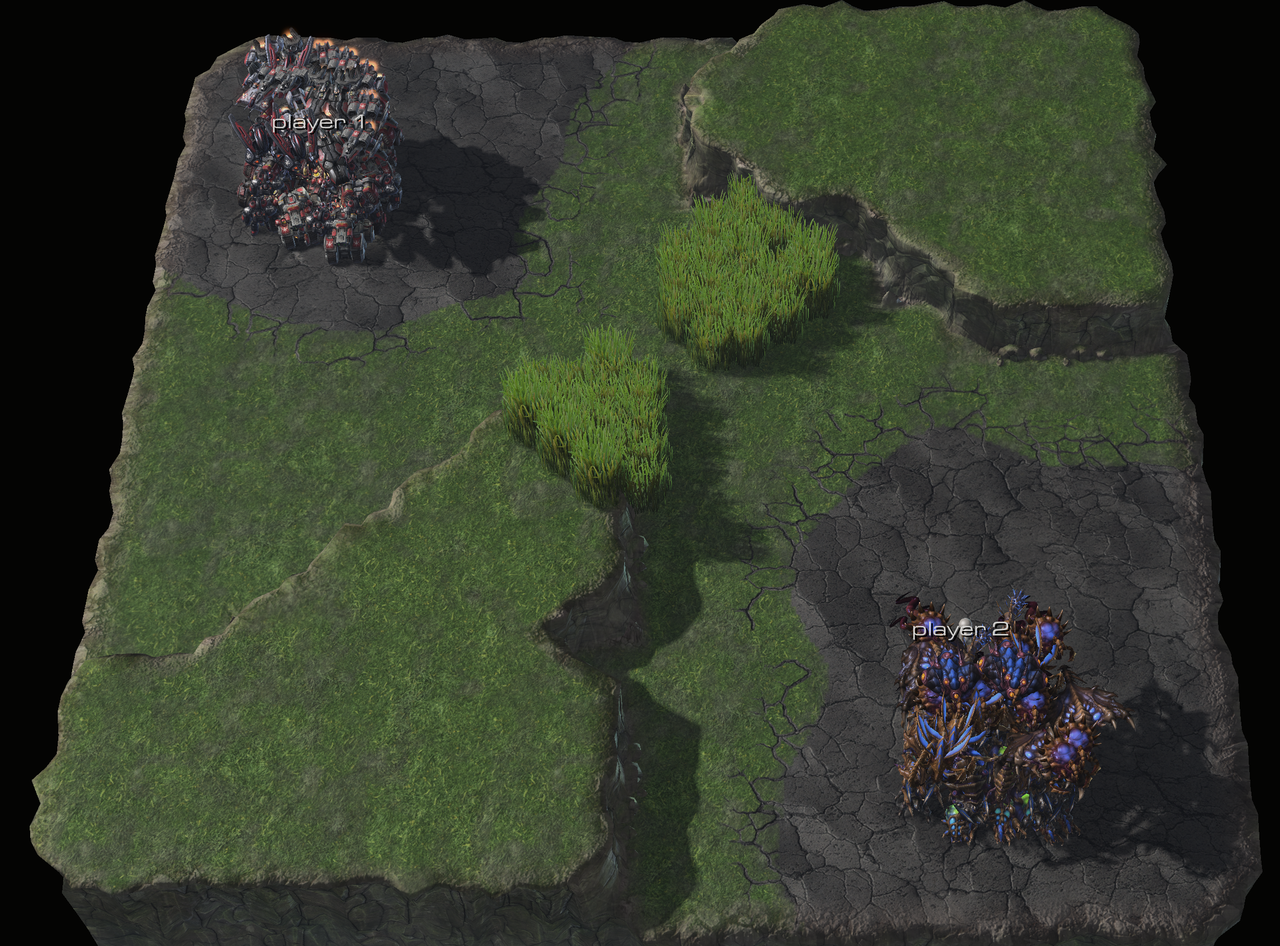}
        \vspace{-0.2cm}
        \caption{Terran vs Zerg on Bush}
        \label{fig:tvz_bush}
    \end{minipage}
\end{figure*}

\begin{figure*}[t]
    \centering
    \begin{minipage}[t]{0.47\textwidth}
        \centering
        \includegraphics[width=0.78\textwidth]{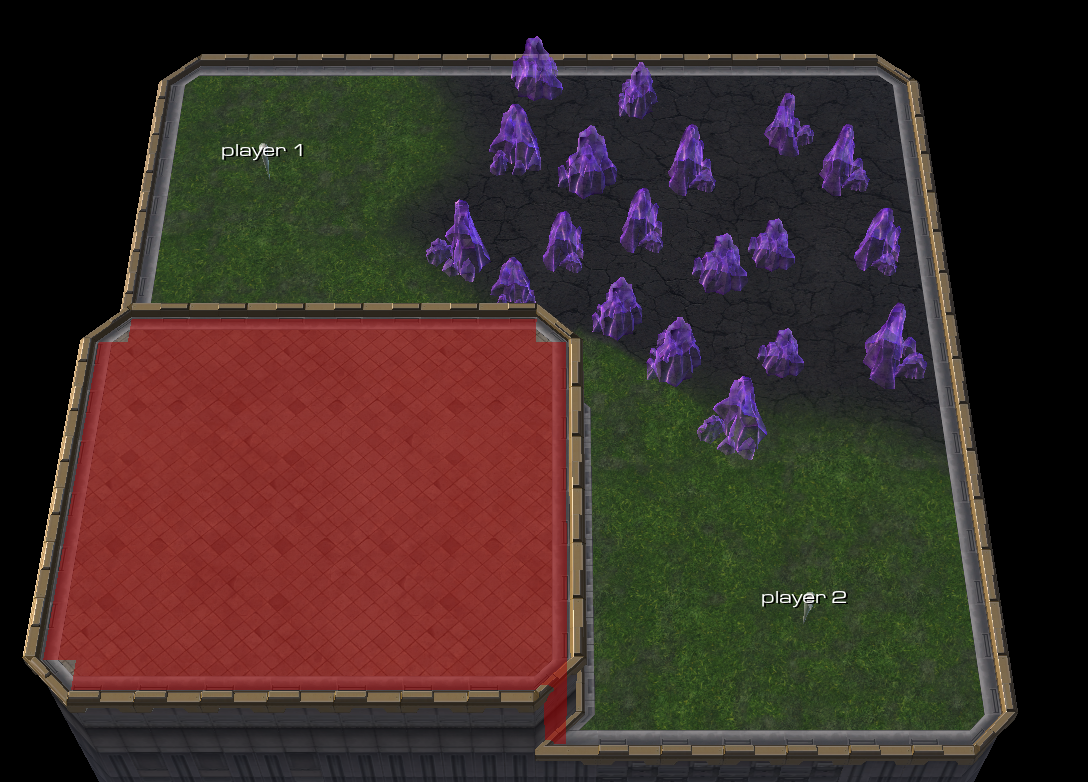}
        \vspace{-0.2cm}
        \caption{Corner map layout}
        \label{fig:corner}
    \end{minipage}
    \hfill
    \begin{minipage}[t]{0.47\textwidth}
        \centering
        \includegraphics[width=0.78\textwidth]{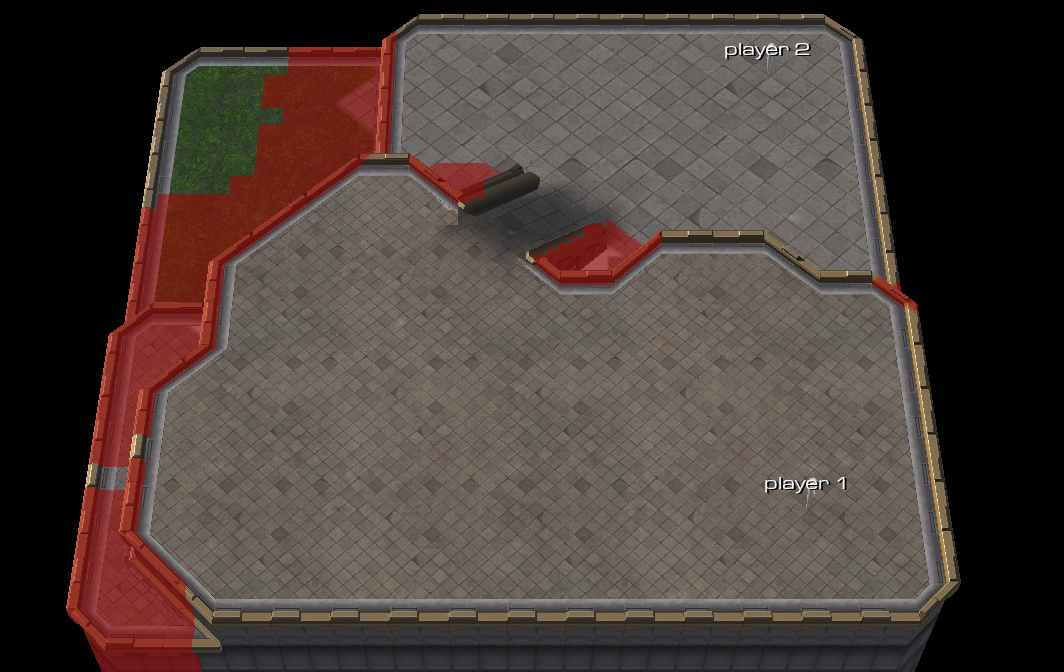}
        \vspace{-0.2cm}
        \caption{Main map layout}
        \label{fig:main}
    \end{minipage}
    
    \vspace{0.12cm}
    
    \begin{minipage}[t]{0.47\textwidth}
        \centering
        \includegraphics[width=0.78\textwidth]{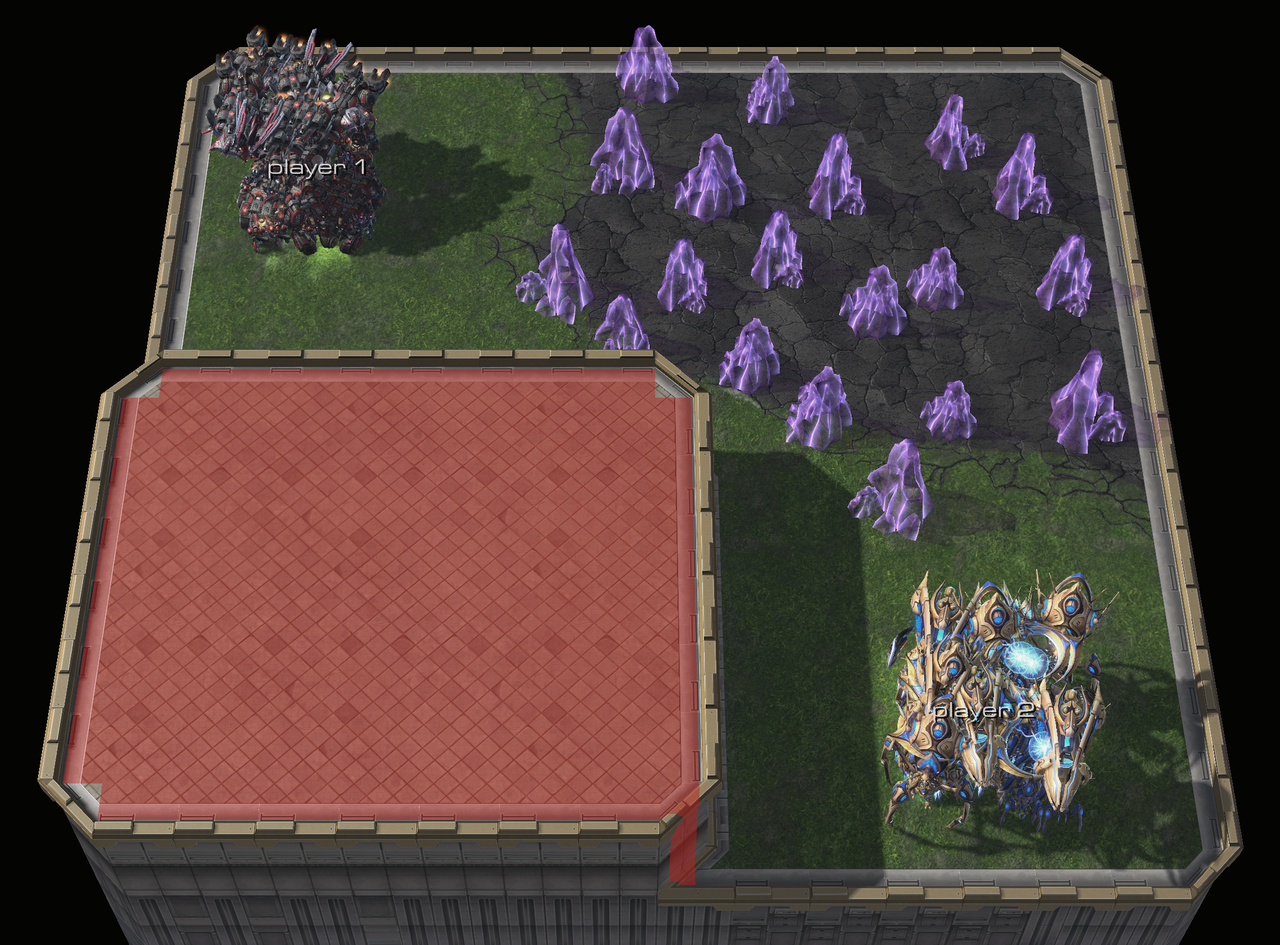}
        \vspace{-0.2cm}
        \caption{Terran vs Protoss on Corner}
        \label{fig:tvp_corner}
    \end{minipage}
    \hfill
    \begin{minipage}[t]{0.47\textwidth}
        \centering
        \includegraphics[width=0.78\textwidth]{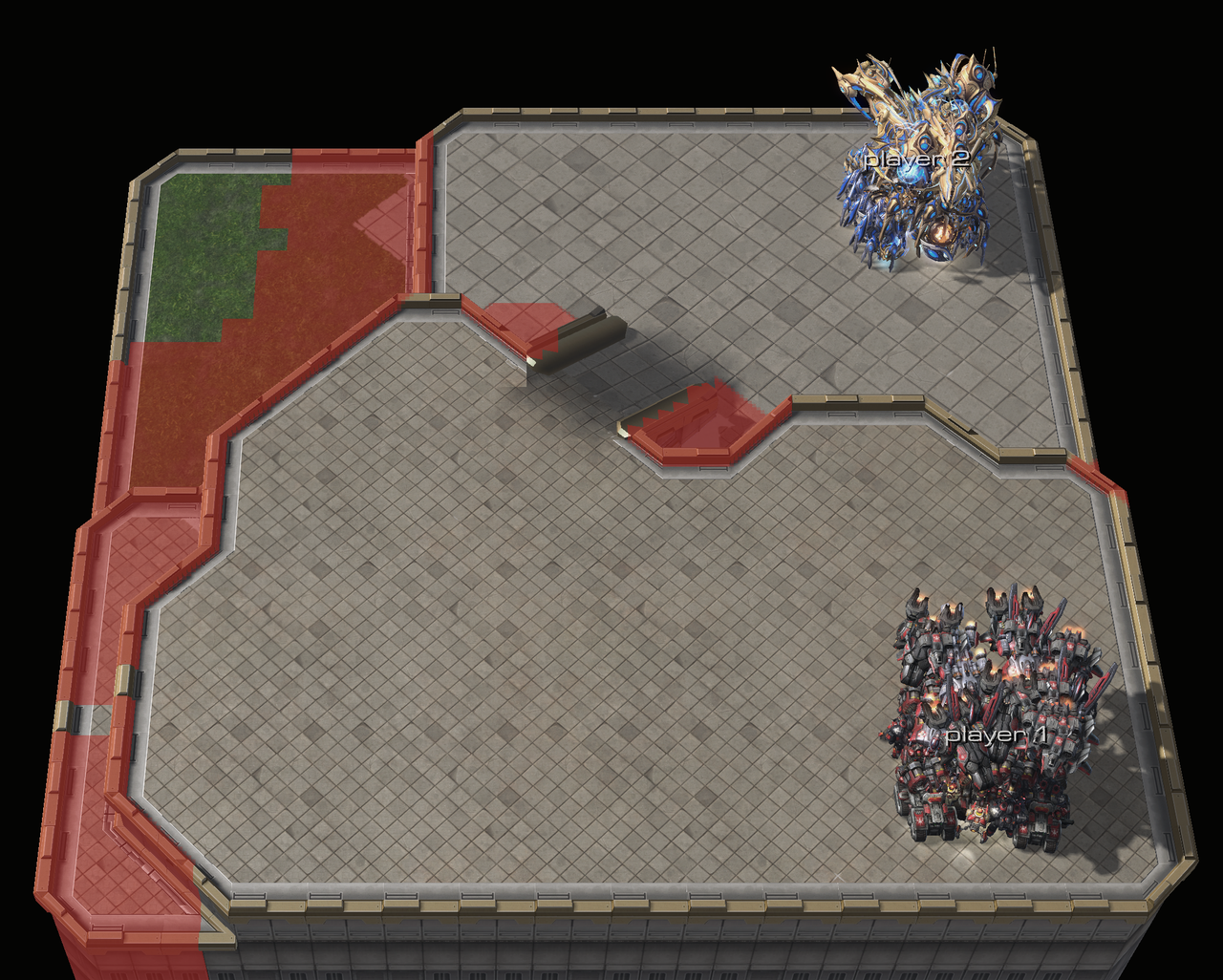}
        \vspace{-0.2cm}
        \caption{Terran vs Protoss on Main}
        \label{fig:tvp_main}
    \end{minipage}
    
    \vspace{0.12cm}
    
    \begin{minipage}[t]{0.47\textwidth}
        \centering
        \includegraphics[width=0.78\textwidth]{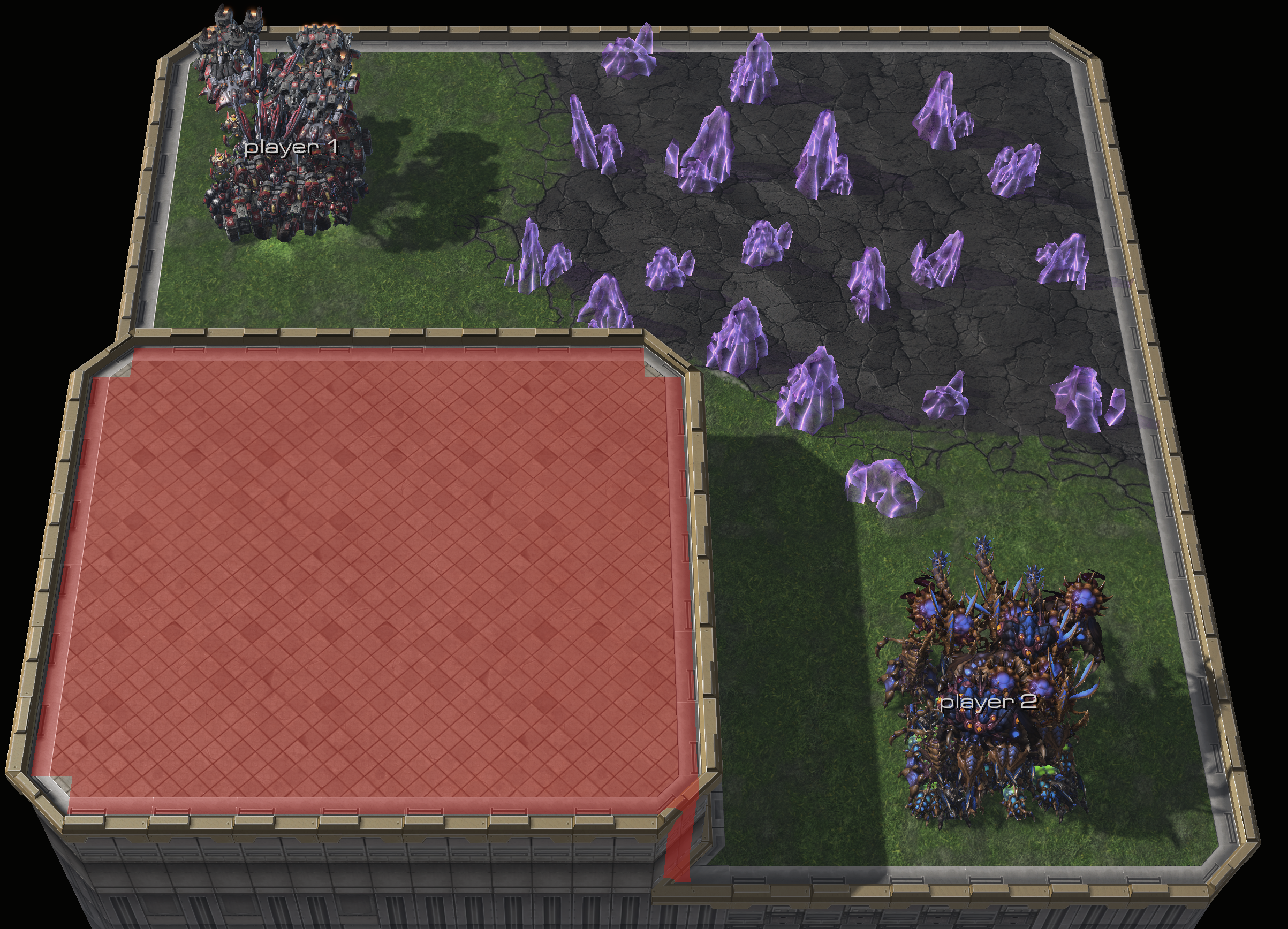}
        \vspace{-0.2cm}
        \caption{Terran vs Zerg on Corner}
        \label{fig:tvz_corner}
    \end{minipage}
    \hfill
    \begin{minipage}[t]{0.47\textwidth}
        \centering
        \includegraphics[width=0.78\textwidth]{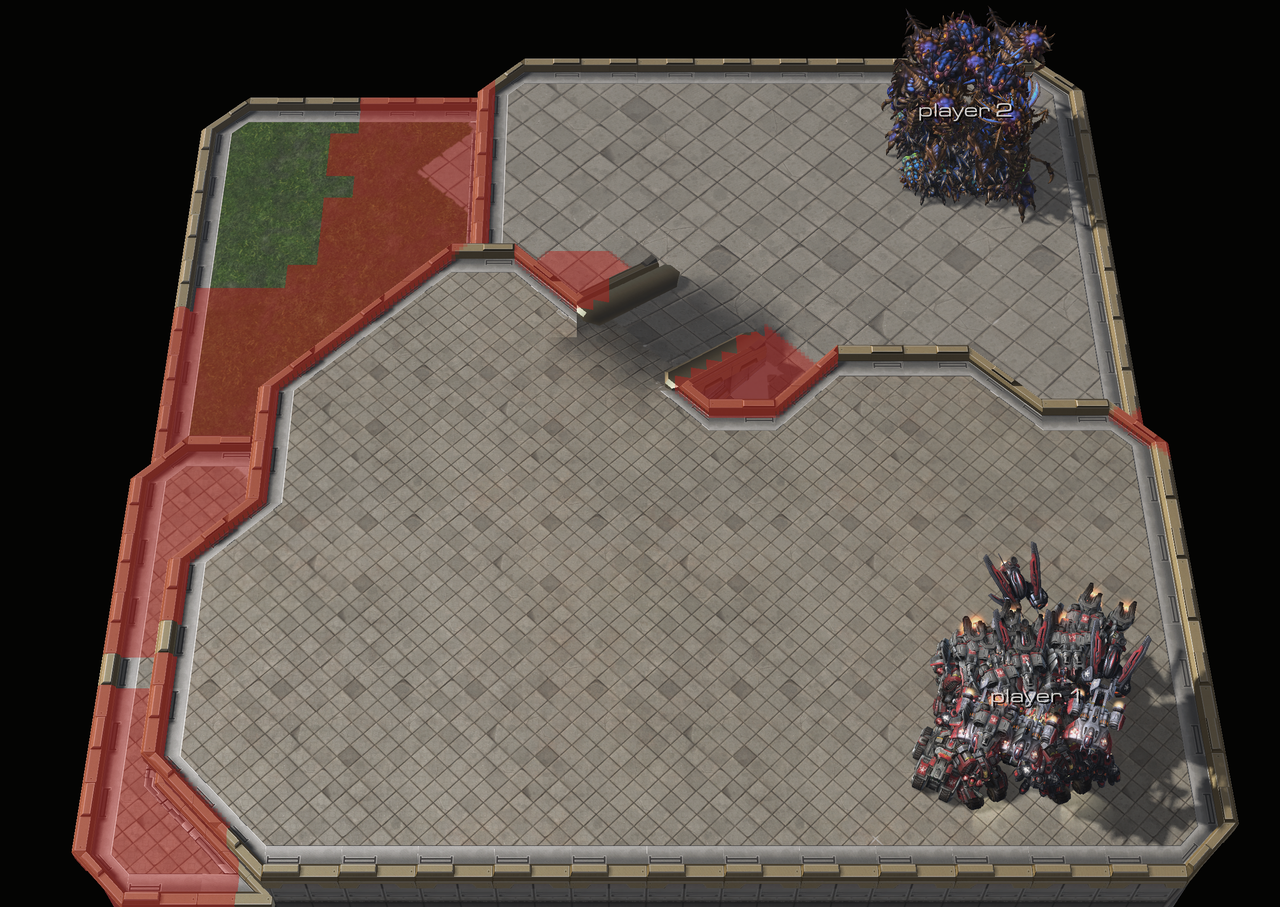}
        \vspace{-0.2cm}
        \caption{Terran vs Zerg on Main}
        \label{fig:tvz_main}
    \end{minipage}
\end{figure*}

\begin{figure*}[t]
    \centering
    \begin{minipage}[t]{0.47\textwidth}
        \centering
        \includegraphics[width=0.78\textwidth, keepaspectratio]{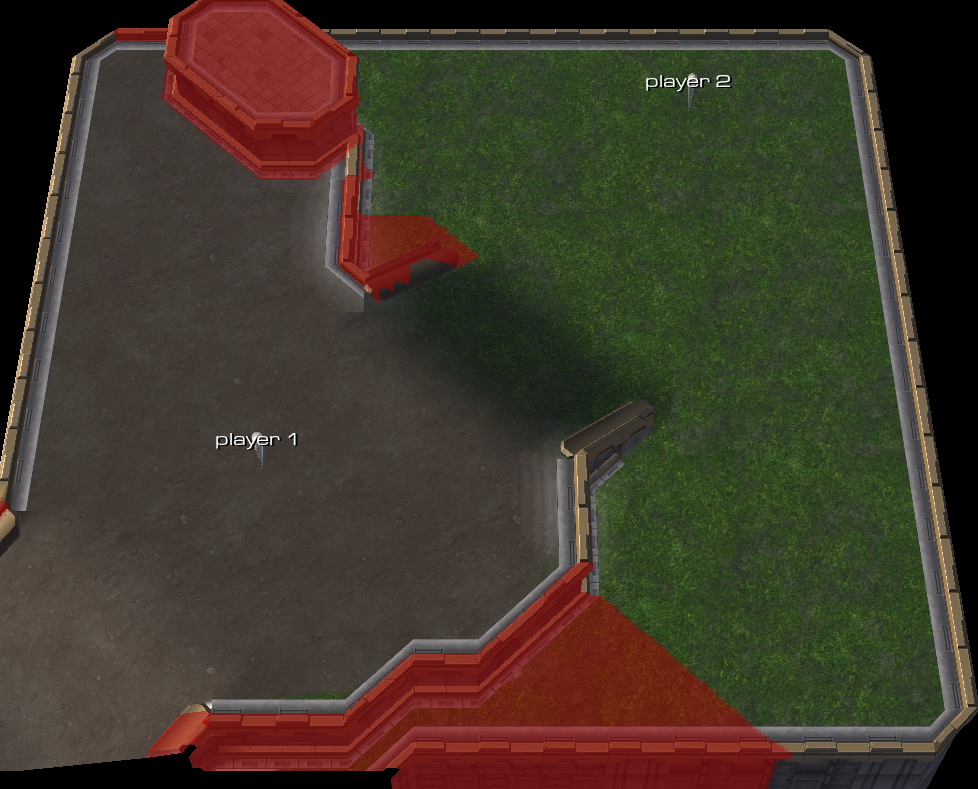}
        \vspace{-0.2cm}
        \caption{Ramp map layout}
        \label{fig:ramp}
    \end{minipage}
    \hfill
    \begin{minipage}[t]{0.47\textwidth}
        \centering
        \includegraphics[width=0.78\textwidth, keepaspectratio]{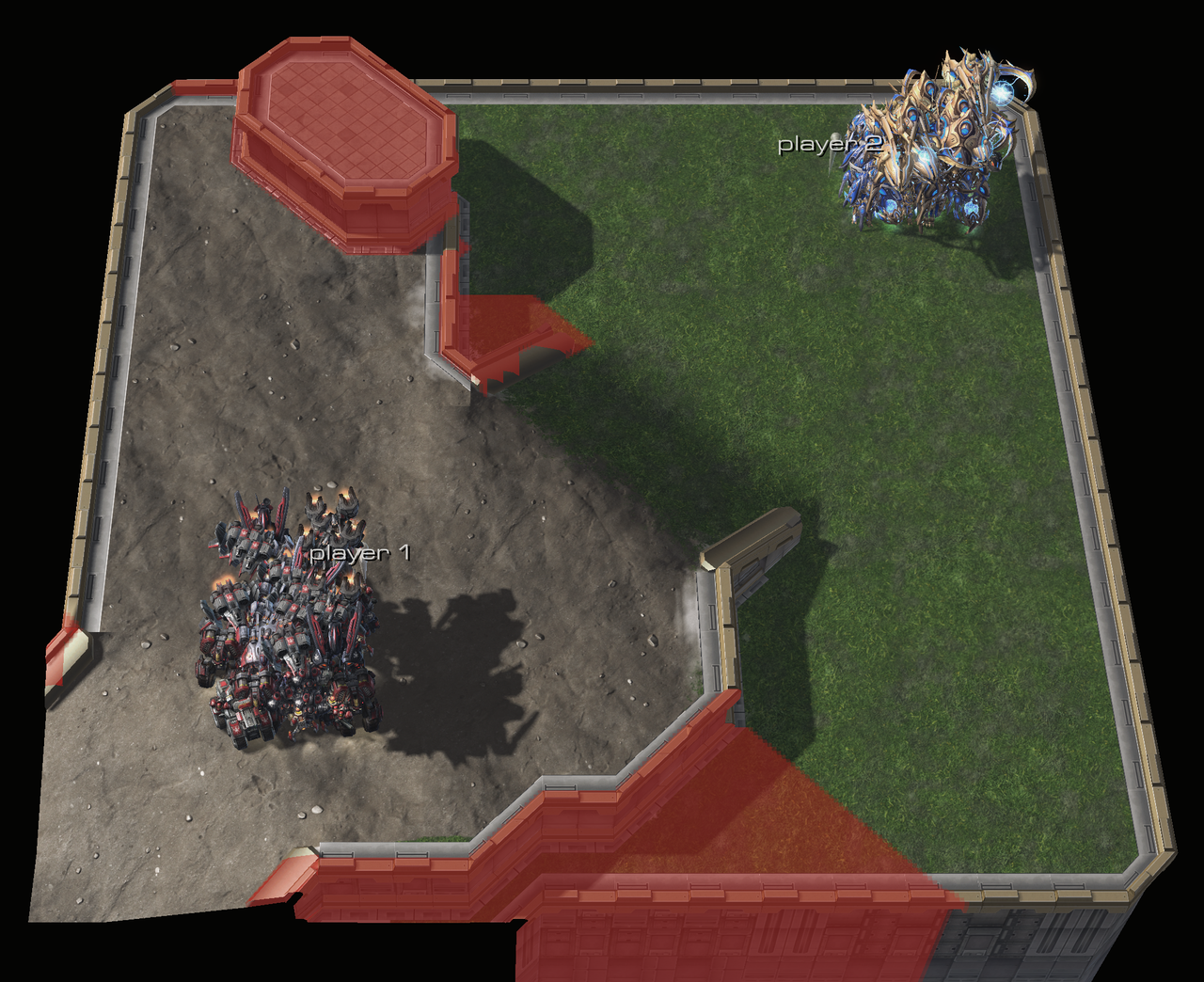}
        \vspace{-0.2cm}
        \caption{Terran vs Protoss on Ramp}
        \label{fig:tvp_ramp}
    \end{minipage}
    
    \vspace{0.12cm}
    
    \begin{minipage}[t]{0.47\textwidth}
        \centering
        \includegraphics[width=0.78\textwidth, keepaspectratio]{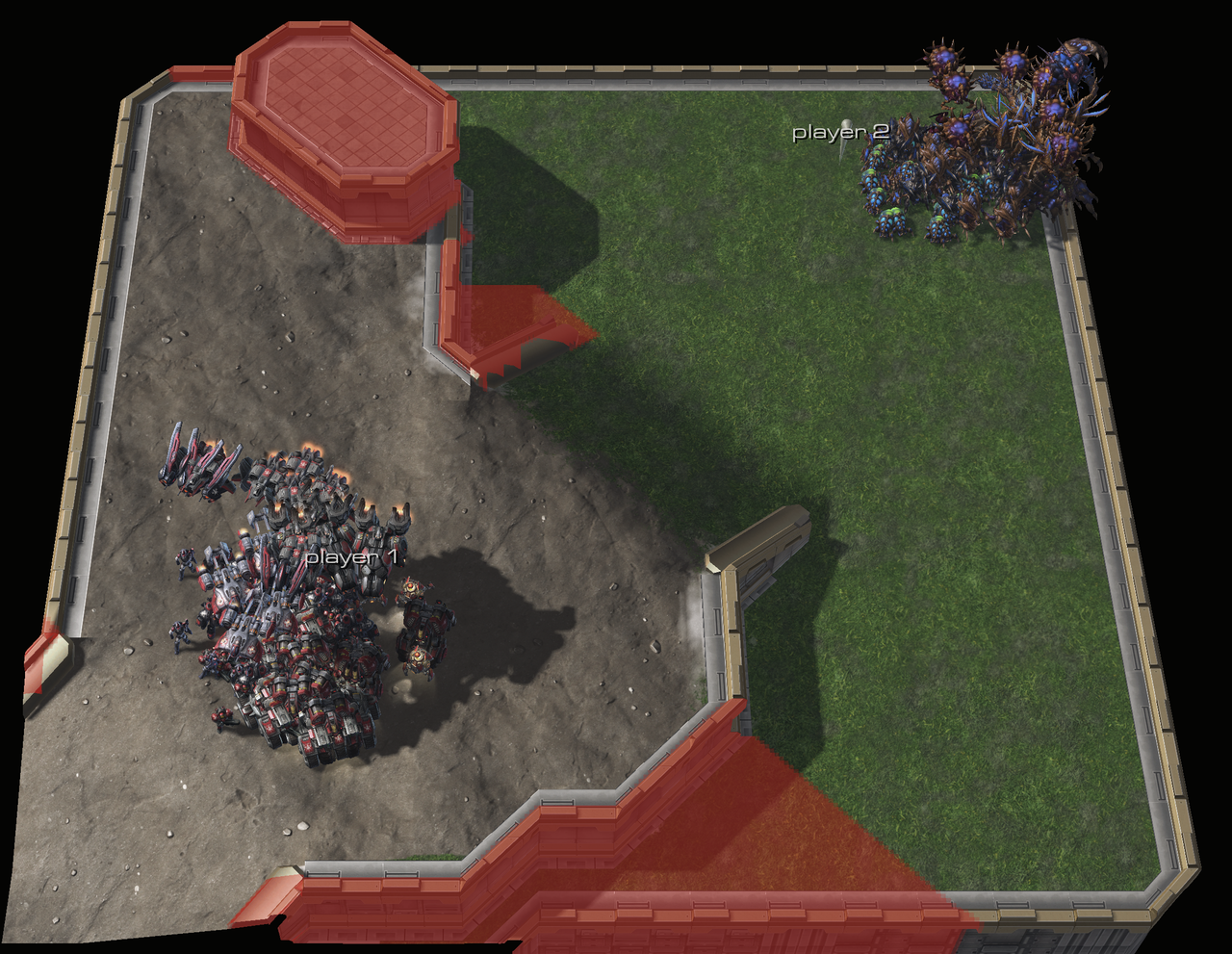}
        \vspace{-0.2cm}
        \caption{Terran vs Zerg on Ramp}
        \label{fig:tvz_ramp}
    \end{minipage}
    \hfill
    \begin{minipage}[t]{0.47\textwidth}
        \centering
    \end{minipage}
\end{figure*}

\FloatBarrier
\section{Overcooked Details}
\label{app:overcooked_training}

\subsection{MARL Training Configuration}

For the Overcooked experiments, we employ the E3T (Efficient End-to-End Training) algorithm~\citep{yan2023efficient} as our MARL training framework. The training configuration is detailed in Table~\ref{tab:overcooked_params}.

\begin{table}[tbp]
\centering
\caption{Overcooked MARL training hyperparameters}
\label{tab:overcooked_params}
\normalsize
\begin{tabular}{@{}ll@{}}
\toprule
\textbf{Parameter} & \textbf{Value} \\
\midrule
Algorithm & E3T \\
Number of agents & 2 \\
Episode length & 400 \\
\makecell[l]{Number of environment steps \\ per curriculum} & $10^7$ \\
PPO epochs & 15 \\
Number of mini-batches & 1 \\
Rollout threads & 100 \\
Evaluation threads & 10 \\
Evaluation interval & 20 episodes \\
\midrule
\multicolumn{2}{@{}l@{}}{\textbf{Entropy Regularization}} \\
Entropy coefficients & [0.2, 0.05, 0.01] \\
\makecell[l]{Entropy coefficient \\ horizons} & [0, $6 \times 10^6$, $10^7$] \\
\midrule
\multicolumn{2}{@{}l@{}}{\textbf{Network Architecture}} \\
CNN layers & \makecell[l]{[(32, $3\times3$, stride=1), \\ (64, $3\times3$, stride=1), \\ (32, $3\times3$, stride=1)]} \\
Recurrent policy & LSTM \\
Shared policy & True \\
\midrule
\multicolumn{2}{@{}l@{}}{\textbf{E3T Specific}} \\
Epsilon (diversity bonus) & 0.25 \\
Weights copy factor & 0.1 \\
Random index & Enabled \\
\midrule
\multicolumn{2}{@{}l@{}}{\textbf{Curriculum-specific}} \\
Reward shaping horizon & $10^8$ \\
Budget per curriculum & $10^7$ timesteps \\
\bottomrule
\end{tabular}
\end{table}

\subsection{Curriculum Design for Overcooked}

The curriculum progression in Overcooked is structured around three key dimensions:

\begin{enumerate}
    \item \textbf{Layout Complexity}: Starting from simple layouts (e.g., \texttt{small\_corridor}) with direct paths between stations, progressing to complex layouts with obstacles and longer navigation requirements.
    
    \item \textbf{Order Complexity}: Beginning with single-ingredient dishes, advancing to multi-ingredient recipes requiring precise coordination between agents.
    
    \item \textbf{Temporal Constraints}: Initially allowing unlimited time for order completion, then introducing time pressure and simultaneous order requirements.
\end{enumerate}

The acceptance criterion $\mathcal{P}(\pi|C) = 1$ is achieved when all required orders are completed within the episode. After meeting this criterion, agents can pursue bonus orders to maximize total deliveries. The entropy regularization schedule ensures exploration early in training (high entropy) while converging to more deterministic policies as training progresses.

For layouts with particular navigation challenges (e.g., \texttt{small\_corridor}), we adjust the entropy coefficient horizons to [0, $8 \times 10^6$, $10^7$] to allow for extended exploration before convergence. The shared policy architecture enables agents to learn cooperative behaviors more efficiently by sharing representations while maintaining individual action distributions.

\section{Overcooked Experimental Results Details}
\label{app:overcooked_results}
This appendix provides detailed experimental results for the Overcooked environment, showing the complete curriculum evolution and performance metrics for two different map configurations.

\subsection{Map Configuration 1}

\subsubsection{Final Layout Specification}
\begin{figure}[tbp]
    \centering
    \includegraphics[width=0.8\linewidth]{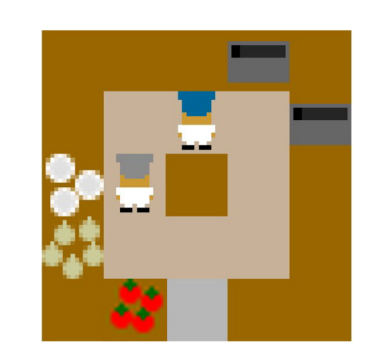}
    \caption{Overcooked Map 1.}
    \label{fig:overcook_map1}
\end{figure}
{\normalsize
\Needspace{18\baselineskip}
\begin{verbatim}
{
  "grid": "XXXPX
          X 2 P
          D1X X
          O   X
          XTSXX", 
  "start_all_orders": [
    {"ingredients": ["onion", "tomato"]}, 
    {"ingredients": ["onion", "onion"]}, 
    {"ingredients": ["onion", "tomato", "tomato"]}, 
    {"ingredients": ["onion", "onion", "tomato"]}, 
    {"ingredients": ["onion", "onion", "onion"]}
  ], 
  "recipe_value": [10, 10, 20, 20, 20], 
  "recipe_time": [10, 10, 20, 20, 20]
}
\end{verbatim}
}

Grid Legend: X=Wall, P=Pot, D=Dish, O=Onion, T=Tomato, S=Service

\subsubsection{Curriculum Evolution Results}
The Overcooked Map 1 evolution path and result details are shown in Table~\ref{tab:overcooked_map1_curriculum}.

\begin{table}[tbp]
\centering
\caption{Map 1: Curriculum progression and performance metrics (A0 Del: Agent0 Delivery; A1 Del: Agent1 Delivery; Total: Total Delivery)}
\label{tab:overcooked_map1_curriculum}
\normalsize
\setlength{\tabcolsep}{3pt} 
\begin{tabular}{@{}lccccc@{}}
\toprule
\textbf{Task} & \textbf{Orders} & \textbf{A0 Del.} & \textbf{A1 Del.} & \textbf{Total} & \textbf{Reward} \\
\midrule
Task 0 & 3 & 11.75 & 12.33 & 24.08 & 239.2 \\
Task 1 & 4 & 14.56 & 14.33 & 28.89 & 288.1 \\
Task 2 (Final) & 5 & 14.74 & 14.36 & 29.10 & 290.9 \\
\midrule
Direct Training & 5 & 11.93 & 12.18 & 24.11 & 240.5 \\
\bottomrule
\end{tabular}
\end{table}

\subsubsection{Detailed Performance Metrics Comparison}
The Overcooked Map 1 result details and metrics are shown in Table~\ref{tab:overcooked_map1_metrics}.

\begin{table}[tbp]
\centering
\caption{Map 1: Key performance indicators for final task (A0: Agent0; A1: Agent1)}
\label{tab:overcooked_map1_metrics}
\normalsize
\setlength{\tabcolsep}{3pt}
\begin{tabular}{@{}lcccccc@{}}
\toprule
 & \multicolumn{2}{c}{\textbf{EvoCurr}} & \multicolumn{2}{c}{\textbf{Direct}} \\
\cmidrule(lr){2-3} \cmidrule(lr){4-5}
\textbf{Metric} & \textbf{A0} & \textbf{A1} & \textbf{A0} & \textbf{A1} \\
\midrule
Onion in Pot & 15.77 & 15.94 & 13.38 & 13.57 \\
Tomato in Pot & 0.31 & 0.32 & 0.21 & 0.26 \\
Dish Pickup & 7.81 & 7.63 & 6.15 & 6.34 \\
Soup Pickup & 7.68 & 7.55 & 6.31 & 6.33 \\
Cook Actions & 7.44 & 8.67 & 6.34 & 7.30 \\
Delivery & 7.37 & 7.19 & 5.99 & 6.11 \\
Idle Move & 9.40 & 8.68 & 14.90 & 15.10 \\
\bottomrule
\end{tabular}
\end{table}

\subsection{Map Configuration 2}

\subsubsection{Final Layout Specification}
\begin{figure}[tbp]
    \centering
    \includegraphics[width=0.8\linewidth]{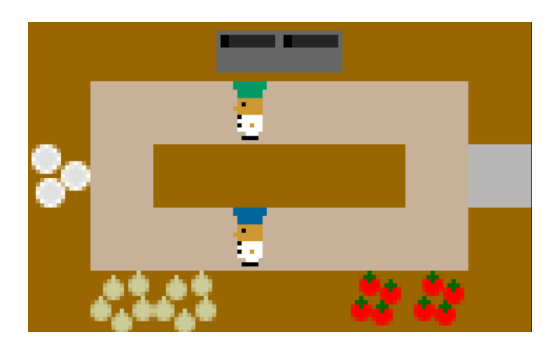}
    \caption{Overcooked Map 2.}
    \label{fig:overcook_map2}
\end{figure}
{\small
\Needspace{18\baselineskip}
\begin{verbatim}
{
  "grid": "XXXPDPXXX
          X   2   X
          X XXXXX X
          X   1   X
          XXXOSTXXX",
  "start_all_orders": [
    {"ingredients": ["onion", "tomato"]},
    {"ingredients": ["onion", "onion"]},
    {"ingredients": ["onion", "tomato", "tomato"]},
    {"ingredients": ["onion", "onion", "tomato"]},
    {"ingredients": ["onion", "onion", "onion"]}
  ],
  "recipe_value": [10, 10, 20, 20, 20],
  "recipe_time": [10, 10, 20, 20, 20]
}
\end{verbatim}
}

Grid Legend: X=Wall, P=Pot, D=Dish, O=Onion, T=Tomato, S=Service

\subsubsection{Curriculum Evolution Results}
The Overcooked Map 2 evolution path and result details are shown in Table~\ref{tab:overcooked_map2_curriculum}. Task 1 includes two single-ingredient recipes alongside complex recipes for an easier transition.

\begin{table}[tbp]
\centering
\caption{Map 2: Curriculum progression and performance metrics (A0 Del: Agent0 Delivery; A1 Del: Agent1 Delivery; Total: Total Delivery)}
\label{tab:overcooked_map2_curriculum}
\small
\setlength{\tabcolsep}{3pt}
\begin{tabular}{@{}lccccc@{}}
\toprule
\textbf{Task} & \textbf{Ord.} & \textbf{A0 Del.} & \textbf{A1 Del.} & \textbf{Total} & \textbf{Reward} \\
\midrule
Task 0 & 3 & 7.28 & 7.25 & 14.53 & 290.3 \\
Task 1 & 5* & 7.32 & 7.28 & 14.60 & 291.6 \\
Task 2 (Final) & 5 & 7.82 & 10.82 & 18.64 & 186.2 \\
\midrule
Direct Training & 5 & 7.40 & 8.96 & 16.36 & 163.4 \\
\bottomrule
\end{tabular}
\end{table}

\subsubsection{Detailed Performance Metrics Comparison}
The Overcooked Map 2 result details and metrics are shown in Table~\ref{tab:overcooked_map2_metrics}.

\begin{table}[tbp]
\centering
\caption{Map 2: Key performance indicators for final task (A0: Agent0; A1: Agent1)}
\label{tab:overcooked_map2_metrics}
\normalsize
\setlength{\tabcolsep}{2.5pt}
\begin{tabular}{@{}lrrrr@{}}
\toprule
 & \multicolumn{2}{c}{\textbf{EvoCurr}} & \multicolumn{2}{c}{\textbf{Direct}} \\
\cmidrule(lr){2-3} \cmidrule(lr){4-5}
\textbf{Metric} & \textbf{A0} & \textbf{A1} & \textbf{A0} & \textbf{A1} \\
\midrule
Onion in Pot & 7.91 & 5.95 & 9.85 & 9.29 \\
Tomato in Pot & 3.17 & 3.87 & 0.08 & 0.07 \\
Dish Pickup & 5.45 & 4.20 & 4.32 & 4.81 \\
Soup Pickup & 4.12 & 5.62 & 4.24 & 4.79 \\
Cook Actions & 5.45 & 4.85 & 5.16 & 4.38 \\
Delivery & 3.92 & 5.42 & 3.71 & 4.49 \\
Size-2 Del. & 3.84 & 5.27 & 3.65 & 4.40 \\
Size-3 Del. & 0.07 & 0.14 & 0.05 & 0.08 \\
Idle Move & 14.42 & 14.21 & 10.49 & 11.28 \\
\bottomrule
\end{tabular}
\end{table}

\section{Claude Code Behavior-Tree Generation Details}
\label{app:claude_code_behavior_tree}
\subsection{Execution Results}
Claude Code is evaluated as an autonomous coding-agent baseline. It
generates its own intermediate curricula and iteratively edits the
behavior-tree implementation using execution feedback.

\begin{table}[!ht]
  \centering
  \setlength{\tabcolsep}{4pt}
  \renewcommand{\arraystretch}{1.15}
  \caption{Claude Code execution outcomes on self-designed curricula and
  the held-out final task.}
  \label{tab:claude_code_execution_summary}
  \begin{tabular}{l c c c}
    \toprule
    \textbf{Task} & \textbf{Attempts} & \textbf{Outcome} & \textbf{Win rate} \\
    \midrule
    Curriculum 1 & 1 & 5W/0D/0L & \textbf{100\%} \\
    Curriculum 2 & 1 & 5W/0D/0L & \textbf{100\%} \\
    Curriculum 3 & 1 & 5W/0D/0L & \textbf{100\%} \\
    Curriculum 4 & 1 & 5W/0D/0L & \textbf{100\%} \\
    Curriculum 5 & 1 & 5W/0D/0L & \textbf{100\%} \\
    \midrule
    \textbf{Final task} & 2 & 0W/1D/9L & \textbf{0\%} \\
    \bottomrule
  \end{tabular}
\end{table}

\begin{flushleft}
\vspace{0.2cm} 

\textbf{Final winning rate: 0\%} \\
\textbf{Total time:} $\sim$90 min \\
\textbf{Evaluation budget:} 100+ simulated matches \\[0.8em]

Despite passing all self-designed intermediate curricula, Claude Code
does not solve the final target task within this budget. Its final
evaluation produces 0 wins, 1 draw, and 9 losses.
\end{flushleft}

\subsection{Curriculum Generation Process}

The following trajectory illustrates the intermediate curricula generated
by Claude Code, ordered from easier to harder settings.

\begin{table}[t]
\centering
\small
\setlength{\tabcolsep}{3pt}
\renewcommand{\arraystretch}{1.12}

\caption{Claude Code self-generated curriculum trajectory.
Unit entries are written as \emph{unit count (upgrade)}.}
\label{tab:claude_code_curricula}

\begin{tabularx}{\columnwidth}{@{}cX@{}}
\toprule
\textbf{Step} & \textbf{Scenario Configuration} \\
\midrule

C1 &
\textbf{Friendly:}
Marine 20 (Stimpack), Marauder 12 (Stimpack),
Medivac 4 (Heal) \\
&
\textbf{Enemy:}
Zealot 6 (Charge), Stalker 4 (BlinkTech)
\\
\midrule

C2 &
\textbf{Friendly:}
Marine 20 (Stimpack), Marauder 12 (Stimpack),
Medivac 4 (Heal) \\
&
\textbf{Enemy:}
Zealot 10 (Charge), Stalker 8 (BlinkTech)
\\
\midrule

C3 &
\textbf{Friendly:}
Marine 20 (Stimpack), Marauder 12 (Stimpack),
Medivac 4 (Heal) \\
&
\textbf{Enemy:}
Zealot 12 (Charge), Stalker 10 (BlinkTech),
Sentry 4 (ForceField)
\\
\midrule

C4 &
\textbf{Friendly:}
Marine 20 (Stimpack), Marauder 12 (Stimpack),
Medivac 4 (Heal), Ghost 4 (PersonalCloaking) \\
&
\textbf{Enemy:}
Zealot 12 (Charge), Stalker 10 (BlinkTech),
HighTemplar 4 (PsiStormTech)
\\
\midrule

C5 &
\textbf{Friendly:}
Marine 20 (Stimpack), Marauder 12 (Stimpack),
Medivac 4 (Heal), Ghost 6 (PersonalCloaking),
SiegeTank 4 (SiegeTech) \\
&
\textbf{Enemy:}
Zealot 15 (Charge), Stalker 12 (Blink),
HighTemplar 6 (PsiStorm),
Colossus 2 (ExtThermalLance)
\\

\bottomrule
\end{tabularx}
\end{table}

\subsection{Prompt Template Used for Coding Agents}

\label{app:prompt_template}
\normalsize
\begin{tcolorbox}[
    title={Prompt Template}, 
    colback=white,      
    colframe=black,    
    boxrule=0.8pt,      
    arc=0pt,            
    fontupper=\normalsize\ttfamily, 
    left=2mm, right=2mm, top=2mm, bottom=2mm 
]

You have a sc2 control task as following:

\textbf{AGENTS:}
\begin{description}[noitemsep,topsep=0pt,parsep=0pt,leftmargin=1em]
    \item[Marine:] 20, (5,25), Stimpack
    \item[Marauder:] 12, (5,25), Stimpack
    \item[Medivac:] 4, (5,25), Heal
    \item[Ghost:] 8, (5,25), PersonalCloaking
    \item[SiegeTank:] 6, (5,25), SiegeTech
    \item[VikingFighter:] 8, (5,25), AssaultMode
    \item[Cyclone:] 8, (5,25), LockOn
    \item[WidowMine:] 8, (5,25), Burrow
    \item[Raven:] 3, (5,25), HunterSeeker
    \item[Liberator:] 4, (5,25), DefenderMode
\end{description}

\vspace{0.5em} 

\textbf{ENEMIES:}
\begin{description}[noitemsep,topsep=0pt,parsep=0pt,leftmargin=1em]
    \item[Zealot:] 15, (25,5), Charge
    \item[Stalker:] 15, (25,5), BlinkTech
    \item[Sentry:] 10, (25,5), ForceField
    \item[HighTemplar:] 9, (25,5), PsiStormTech
    \item[Colossus:] 4, (25,5), ExtendedThermalLance
    \item[Tempest:] 5, (25,5), GroundAttack
    \item[Disruptor:] 4, (25,5), PurificationNova
    \item[Carrier:] 4, (25,5), InterceptorLaunch
\end{description}

\vspace{0.5em}

You should design the curricula \textbf{from easy to hard} to make sure the final control code can win the final task.

You can change environments by assigning the scenario\_type and map\_name in rollout\_config.py.

Generate the control code and test it in res-temp.py. Prefer changing only the BattleBot class. Use the evaluation feedback to revise the policy.

When you pass the current curriculum with a high winning rate, you can design the next curriculum.
\end{tcolorbox}

\subsection{Generated Code Sample}
\Needspace{18\baselineskip}
\begin{lstlisting}[
    style=mystyle,
    language=Python,
    basicstyle=\ttfamily\scriptsize,
    numbers=left,
    breaklines=true,
    frame=lines,
    framesep=2mm,
    xleftmargin=2em,
    numbersep=4pt
]
class BattleBot(BotAI):
    """
    Terran Auto-Micro Bot - Optimized Version
    Strategy: Immediate Stimpack + Focus Fire Nearest Enemy + Simple Kiting + Medivac Healing
    """

    def __init__(self):
        super().__init__()
        self.stim_used = set()
        self.started = False

    async def on_step(self, iteration: int):
        # Get units
        marines = self.units(UnitTypeId.MARINE)
        marauders = self.units(UnitTypeId.MARAUDER)
        medivacs = self.units(UnitTypeId.MEDIVAC)
        ghosts = self.units(UnitTypeId.GHOST)
        tanks = self.units(UnitTypeId.SIEGETANK) | self.units(UnitTypeId.SIEGETANKSIEGED)

        bio_units = marines | marauders
        all_combat = bio_units | ghosts | tanks
        enemies = self.enemy_units
        enemy_pos = Point2((27, 27))

        if not all_combat.exists:
            return

        # No enemies - quick attack
        if not enemies.exists:
            for unit in all_combat:
                unit.attack(enemy_pos)
            for medivac in medivacs:
                medivac.move(enemy_pos)
            return

        # === Optimization 1: Stimpack immediately upon first enemy sight ===
        if not self.started:
            self.started = True
            for unit in bio_units:
                if unit.health > 20:
                    if unit.type_id == UnitTypeId.MARINE:
                        unit(AbilityId.EFFECT_STIM_MARINE)
                    elif unit.type_id == UnitTypeId.MARAUDER:
                        unit(AbilityId.EFFECT_STIM_MARAUDER)
                    self.stim_used.add(unit.tag)

        # === Optimization 2: Simplified target selection ===
        low_health = enemies.filter(lambda u: u.health_percentage < 0.4)
        if low_health.exists:
            target = low_health.closest_to(bio_units.center)
        else:
            target = enemies.closest_to(bio_units.center)

        # === Optimization 3: Kiting tactics ===
        for unit in bio_units:
            closest = enemies.closest_to(unit.position)
            dist = unit.distance_to(closest)
            # Maintain distance from melee units
            if closest.type_id in [UnitTypeId.ZEALOT, UnitTypeId.ZERGLING]:
                if dist < 3:
                    retreat_pos = unit.position.towards(closest.position, -4)
                    unit.move(retreat_pos)
                    continue
            unit.attack(target)

        # === Ghost EMP ===
        for ghost in ghosts:
            shielded = enemies.filter(lambda u: u.shield > 20)
            if shielded.exists and ghost.energy >= 75:
                emp_target = max(shielded, key=lambda u: enemies.closer_than(3, u).amount)
                ghost(AbilityId.EMP_EMP, emp_target.position)
            else:
                ghost.attack(target)

        # === Tank Control ===
        for tank in tanks:
            if tank.type_id == UnitTypeId.SIEGETANK:
                closest = enemies.closest_to(tank.position)
                dist = tank.distance_to(closest)
                if 3 < dist < 13:
                    tank(AbilityId.SIEGEMODE_SIEGEMODE)
                else:
                    tank.attack(target)
            else:  # SIEGETANKSIEGED
                if tank.distance_to(enemies.closest_to(tank)) < 3:
                    tank(AbilityId.UNSIEGE_UNSIEGE)
                else:
                    tank.attack(target)

        # === Medivac Healing ===
        for medivac in medivacs:
            if bio_units.exists:
                injured = bio_units.filter(lambda u: u.health_percentage < 0.8)
                if injured.exists:
                    heal_target = min(injured, key=lambda u: u.health)
                    medivac(AbilityId.MEDIVACHEAL_HEAL, heal_target)
                else:
                    medivac.move(bio_units.center)
\end{lstlisting}

\FloatBarrier
\section{Computational Experiments}
\label{app:computational_experiments}
We measure the approximate API cost and wall-clock time required to solve a final SC2 task such as Bush (TvP). These measurements use closed-source model APIs and include all curriculum stages needed to reach the final task.

For DeepSeek-V3.1, completing one final task costs approximately \$0.2. For GPT-5 and Claude-4, the cost is approximately \$1.5.

Each individual curriculum stage requires approximately 5--30 minutes. A final task typically requires 4--6 sequential stages, for a total wall-clock time of approximately 1--3 hours.

\end{document}